\documentclass[10pt,journal,compsoc]{IEEEtran}
\usepackage{graphicx}
\usepackage{subcaption}
\usepackage[table]{xcolor}% http://ctan.org/pkg/xcolor
\usepackage{url}
\usepackage[boxed]{algorithm2e}
\usepackage[noend]{algpseudocode}
\usepackage{xspace}
\usepackage{booktabs}
\usepackage{multirow}
\usepackage{siunitx}
\usepackage{amsmath,amssymb}
\usepackage{mathrsfs}
\usepackage{mathtools}
\usepackage{ragged2e}
\usepackage{wrapfig}
\usepackage{color,soul}

\newcommand{\hlcyan}[1]{{#1}}

\usepackage[pagebackref=true,breaklinks=true,letterpaper=true,colorlinks,bookmarks=false]{hyperref}

\ifCLASSOPTIONcompsoc
  \usepackage[nocompress]{cite}
\else
  \usepackage{cite}
\fi
\ifCLASSINFOpdf
\else
\fi
\begin{document}
\title{Training Faster by Separating Modes of Variation in Batch-normalized Models}

\author{Mahdi~M.~Kalayeh,~\IEEEmembership{Member,~IEEE,}
        and~Mubarak~Shah,~\IEEEmembership{Fellow,~IEEE}% <-this % stops a space
\IEEEcompsocitemizethanks{\IEEEcompsocthanksitem M. M. Kalayeh and M. Shah are with the Center for Research in Computer Vision (CRCV), Department
of Computer Science, University of Central Florida, Orlando,
FL, 32816.\protect\\
E-mails: mahdi@eecs.ucf.edu, shah@crcv.ucf.edu
}% <-this % stops an unwanted space
\thanks{}}

\IEEEtitleabstractindextext{%
\begin{abstract}
\justifying
Batch Normalization (BN) is essential to effectively train state-of-the-art deep Convolutional Neural Networks (CNN). It normalizes the layer outputs during training using the statistics of each mini-batch. BN accelerates training procedure by allowing to safely utilize large learning rates and alleviates the need for careful initialization of the parameters. In this work, we study BN from the viewpoint of Fisher kernels that arise from generative probability models. We show that assuming samples within a mini-batch are from the same probability density function, then BN is identical to the Fisher vector of a Gaussian distribution. That means batch normalizing transform can be explained in terms of kernels that naturally emerge from the probability density function that models the generative process of the underlying data distribution. Consequently, it promises higher discrimination power for the batch-normalized mini-batch. However, given the rectifying non-linearities employed in CNN architectures, distribution of the layer outputs show an asymmetric characteristic. Therefore, in order for BN to fully benefit from the aforementioned properties, we propose approximating underlying data distribution not with one, but a mixture of Gaussian densities. Deriving Fisher vector for a Gaussian Mixture Model (GMM), reveals that batch normalization can be improved by independently normalizing with respect to the statistics of disentangled sub-populations. We refer to our proposed soft piecewise version of batch normalization as Mixture Normalization (MN). Through extensive set of experiments on CIFAR-10 and CIFAR-100, using both a 5-layers deep CNN and modern Inception-V3 architecture, we show that mixture normalization reduces required number of gradient updates to reach the maximum test accuracy of the batch normalized model by $\sim$31\%-47\% across a variety of training scenarios. Replacing even a few BN modules with MN in the 48-layers deep Inception-V3 architecture is sufficient to not only obtain considerable training acceleration but also better final test accuracy. \hlcyan{We show that similar observations are valid for 40 and 100-layers deep DenseNet architectures as well}. We complement our study by evaluating the application of mixture normalization to the Generative Adversarial Networks (GANs), where ``mode collapse'' hinders the training process. We solely replace a few batch normalization layers in the generator with our proposed mixture normalization. Our experiments using Deep Convolutional GAN (DCGAN) on CIFAR-10 show that mixture normalized DCGAN not only provides an acceleration of $\sim$58\% but also reaches lower (better) ``Fr\'echet Inception Distance'' (FID) of 33.35 compared to 37.56 of its batch normalized counterpart. 

\end{abstract}

}

\maketitle

\IEEEdisplaynontitleabstractindextext

% \fi
\IEEEpeerreviewmaketitle

\IEEEraisesectionheading{\section{Introduction}\label{sec:introduction}}
% \IEEEPARstart{I}{n} recent years, state-of-the-art performance in a variety of computer vision tasks has continuously improved. In addition to leveraging huge number of training data, most of these progresses were mainly achieved by novel designs such as Inception \cite{szegedy2015going} architecture, residual learning \cite{he2016deep} or dense connections\cite{huang2017densely}.

\IEEEPARstart{I}{n} the context of deep neural networks, the distribution of inputs to each layer is not only heavily dependent on the the previous layers but it also changes as the the network evolves through the training procedure. Ioffe and Szegedy \cite{ioffe2015batch} referred to this phenomenon as \textit{internal covariate shift}. Another source of variation comes from the family of optimization techniques, specifically Stochastic Gradient Descent (SGD), that is used to train the deep neural networks. Although we can normalize a mini-batch at the input to the network, it will not remain normalized after multiple rounds of non-linear operations as we proceed from early layers to deeper ones. Batch Normalization (BN) \cite{ioffe2015batch} was proposed to alleviate the \textit{internal covariate shift} by normalizing layer outputs for each training mini-batch with respect to its very own statistics, specifically mean and variance. 

Batch normalization has been successful in dramatically accelerating training procedure of deep neural networks and is now a standard component in almost all deep convolutional architectures. In fact, it is hard to imagine the possibility of effectively training state-of-the-art architectures such as Inception \cite{szegedy2016rethinking}\cite{szegedy2017inception}, Residual \cite{he2016deep} and Densely connected \cite{huang2017densely} networks, with hundreds of layers, without employing batch normalization.

Batch normalization \cite{ioffe2015batch} and its few extensions \cite{ba2016layer}\cite{ulyanov2017improved}\cite{wu2018group}\cite{ren2016normalizing}, follow a general form (ref. Equation \ref{eq:general_form_normalization}) to normalize a mini-batch, yet differ in the construction of the population over which mini-batch statistics are computed. In this paper, we provide a fresh view on the aforementioned general form of normalization, demonstrating its strong relation to the natural kernels that arise from generative probability models. Specifically, we show that assuming samples within a mini-batch are from the same probability density function, the general form of normalization is identical to the Fisher vector of a Gaussian distribution. Jaakkola and Haussler \cite{jaakkola1999exploiting} proved that given classification labels as latent variables, Fisher kernel is asymptotically never inferior to the maximum \textit{a posteriori} (MAP) decision rule. Therefore, the general form of normalization used in BN and its extensions, not only naturally emerges from the probability density function that models the generative process of the underlying data distribution, but also improves the discrimination power of the representation\footnote{\hlcyan{This is established on the basis of the discriminative derivations of Fisher kernel (ref. Theorem 1 of \cite{jaakkola1999exploiting}). The mathematical derivations can be found in Appendix A of the longer version of \cite{jaakkola1999exploiting}, available at \url{https://people.csail.mit.edu/tommi/papers/gendisc.ps}.}}.

However, for BN \cite{ioffe2015batch} to truly enjoy theses properties, its input activations should follow a Gaussian distribution. We argue that due to the rectifying non-linearities employed in deep neural networks, it is unlikely that such condition is fully satisfied. Specifically, it is hard to believe that, in expectation, a linear combination (convolution operation) of multiple (different channels) distributions with semi-infinite support (output of previous \textit{e.g} ReLU \cite{nair2010rectified}), with at least two modes of variation one for the rectified values mapped to zero and one for the positive values, results in a single Gaussian distribution. We visualize internal activations of both a shallow and very deep architecture (ref. Section \ref{sec:experiments}), and observe that indeed corresponding activations very often illustrate asymmetric characteristic and are better approximated by a mixture model rather than a single Gaussian distribution. This observation builds the core of our work. 

To equip batch normalization with characteristics which Fisher kernel promises, we must first properly approximate the probability density function of the internal representations.
From \cite{titterington1985statistical}, we know that any continuous distribution can be approximated with arbitrary precision using a Gaussian Mixture Model (GMM). Hence, instead  of  computing  one  set  of  statistical measures from the entire population (of instances in the mini-batch) as BN does, we propose normalization on sub-populations which can be identified by disentangling modes of the distribution, estimated via GMM. We refer to our proposed technique as Mixture Normalization (MN). 

While BN can only scale and/or shift the \textit{whole} underlying probability density function, mixture normalization operates like a soft piecewise normalizing transform, capable of completely re-structuring the data distribution by independently scaling and/or shifting individual modes of distribution (ref. Figure \ref{fig:teaser}).

We show that mixture normalization can effectively accelerate its batch normalization \cite{ioffe2017batch} counterpart, through reducing required number of gradient updates to reach the maximum test accuracy of the batch normalized model by $\sim$\textbf{31}\%-\textbf{47}\%, across a variety of training scenarios on CIFAR-10 and CIFAR-100. Mixture normalization handles training in large learning rate regime considerably better than batch normalization, and in majority of cases reaches even a better final test accuracy. It is worth pointing out that such acceleration in training procedure is obtained by solely replacing \textit{a few} batch normalization modules with mixture normalization along the depth of the architecture. This is true both for shallow and very deep and modern architectures such as Inception-V3 \cite{szegedy2016rethinking} and DenseNet \cite{huang2017densely}. 

In summary, the main contributions of this work are as follows:
\begin{itemize}
    \item We demonstrate how normalizing transform, employed in batch normalization and its variants, is related to the kernels from generative probability models.
    \item We show that, the distribution of activations associated with internal layers of deep convolutional neural networks, illustrates an asymmetric characteristic and is very likely to be better estimated using a mixture model.
    \item We propose Mixture Normalization (MN), where a Gaussian mixture model initially identifies modes of variation in the underlying distribution, and then each sample in the mini-batch is normalized using mean and standard deviation of the mixture component to which it belongs to.
    \item We confirm through extensive set of experiments on CIFAR-10 and CIFAR-100, that Mixture Normalization not only significantly accelerates convergence of batch normalized models but also achieves better final test accuracy.
\end{itemize}

The remainder of this paper is organized as follows. Section \ref{sec:related_work} offers a detailed review of batch normalization and its various extensions. We then demonstrate the relationship between batch normalization and kernels from generative probability models in Section \ref{sec:generative_probability_models}. This lays the groundwork for presenting our proposed mixture normalization in Section \ref{sec:mixture_normalization}. Experimental results are shown in Section \ref{sec:experiments}. Section \ref{sec:discussion} provides computational complexity analysis along with multiple detailed discussions attempting at an in-depth behavioral analysis of the mixture normalization. Finally, we conclude the paper in Section \ref{sec:conclusion}.

\section{Related Work}\label{sec:related_work}
\subsection{Batch Normalization}\label{sec:batch_normalization}
Let's consider $x\in\mathop{\mathbb{R}^{N\times C\times H\times W}}$, a 4-D activation tensor in a convolutional neural network where $N$, $C$, $H$ and $W$ are respectively the batch, channel, height and width axes. BN \cite{ioffe2015batch} computes the mini-batch mean ($\mu_{\mathcal{B}}$) and standard deviation ($\sigma_{\mathcal{B}}$), formulated in Equation \ref{eq:batch_normalization1}, over the set $\mathcal{B}=\{x_{1\dots m}:m\in{[1,N]\times[1,H]\times[1,W]}\}$, where $x$ is flattened across all but channel axis, and $\epsilon$ is used for numerical stability.

\begin{equation}\label{eq:batch_normalization1}
\mu_{\mathcal{B}} = \frac{1}{m}\mathop{\mathbb{\sum}}_{i=1}^{m}x_{i}\quad
\sigma_{\mathcal{B}} = \sqrt{\frac{1}{m}\mathop{\mathbb{\sum}}_{i=1}^{m}(x_{i}-\mu_{\mathcal{B}})^2+\epsilon}
\end{equation}

If we assume that samples within the mini-batch are from the same distribution, the transform $x\to\hat{x}$ shown in Equation \ref{eq:batch_normalization2}, generates a zero mean and unit variance distribution. Then, BN uses learnable scale ($\gamma$) and shift ($\beta$) parameters to transform the normalized distribution into one with $\beta$ mean and $\gamma$ standard deviation.

\begin{equation}\label{eq:batch_normalization2}
\hat{x}_{i} = \frac{1}{\sigma_{\mathcal{B}}}(x_{i} - \mu_{\mathcal{B}})\quad y_{i} = \gamma\hat{x}_{i} + \beta  
\end{equation}

Following Ioffe and Szegedy's \cite{ioffe2015batch} terminology, we refer to the transform
\begin{equation}
\text{BN}\textsubscript{$\gamma,\beta$}: x_{1\dots m} \to y_{1\dots m}
\end{equation}
as the \textit{Batch Normalizing Transform}. In most of its applications, BN  normalizes the output of a convolution layer just before the non-linear activation function (\textit{e.g} ReLU \cite{nair2010rectified}), which separates two consecutive convolution layers. As we discussed before, normalized activations, $\hat{x}_{1\dots m}$, under the assumption that $x_{1\dots m}$ are from the same distribution, are of zero mean and unit variance. Hence, applying ReLU function to $\hat{x}_{1\dots m}$ can be approximately seen as rectifying half of the distribution. This may not necessarily be the optimum case from the perspective of minimizing the objective function during training. That is why the \textit{learnable} scale ($\gamma$) and more crucially shift ($\beta$) parameter are important, as they allow the training procedure to shape the behavior of the non-linearity and consequently the entire model. It is easy to see (ref. Equation \ref{eq:scale_shift_behavior}) that BN enables the the model to alternate between two extreme cases of ignoring the non-linearity and completely clipping the activations. While in the former, the effective depth of the network reduces as two back-to-back linear operations can be seen as one, the latter prevents the activations from propagating to the next layer.
\begin{equation}\label{eq:scale_shift_behavior}
\begin{split}
        \big\{\forall\gamma\exists\eta\lim\limits_{\beta\to+\eta}\mathop{\text{ReLU}\big(\text{BN}_{\gamma,\beta}(x)\big)}=\text{BN}_{\gamma,\beta}(x)\vert 0<\gamma,\eta\ll\infty\big\},\\
        \big\{\forall\gamma\exists\eta\lim\limits_{\beta\to-\eta}\mathop{\text{ReLU}\big(\text{BN}_{\gamma,\beta}(x)\big)}=0\vert 0<\gamma,\eta\ll\infty\big\}.\\
\end{split}
\end{equation}

The behavior of BN at inference is slightly different from training phase. To cope with potential discrepancy between distributions and dependency of each normalized activation to other instances in the mini-batch, BN accumulates a running average of the statistics at the training phase and uses them for normalizing mini-batches at inference. Hence, unlike training, where each mini-batch is normalized with respect to its very own statistics, all the mini-batches at inference use the same running average statistics for normalization while the scale and shift parameters are frozen. This workaround is effective when the size of the mini-batch is large, its instances are i.i.d. samples from training distribution and pre-computed statistics do not change \cite{ioffe2017batch}. However, in the absence of these conditions, estimation of mean and variance becomes less accurate at mini-batch level, hence affecting the running average statistics, and consequently results in performance degradation. To address these drawbacks, Ioffe \cite{ioffe2017batch} proposed \textit{Batch Renormalization} where a per-dimension affine transformation is applied to the normalized activations as 
\begin{equation}\label{eq:batch_renormalization}
    \frac{x_{i}-\mu}{\sigma} = \frac{x_{i}-\mu_{\mathcal{B}}}{\sigma_{\mathcal{B}}}\cdot r + d,\,\text{where}\,r=\frac{\sigma_{\mathcal{B}}}{\sigma},\, d=\frac{\mu_{\mathcal{B}}-\mu}{\sigma}.
\end{equation}

Note that the parameters of the transformation, $r$ and $d$, are not trainable, instead they, in expectation, compensate for the difference between per mini-batch and over-population statistics. If the expected values of the per mini-batch statistics match the moving average ones, then the affine transformation reduces to an Identity resulting in batch renormalization \cite{ioffe2017batch} to behave identical to the batch normalization \cite{ioffe2015batch}.

\hlcyan{Large batch training of the neural networks is very well motivated as it reduces the training time by effectively leveraging the parallel and/or distributed computation. Yet the common practices and previous empirical observations \mbox{\cite{keskar2016large}} used to suggest that large batch training would not generalize as well as small batch training. Recently, Hoffer \textit{et. al} \mbox{\cite{hoffer2017train}} showed that there is no inherent generalization gap associated with large batch training once the number of iterations and learning rate are properly adapted. However, they noticed that the dependency of statistics used by batch normalization \mbox{\cite{ioffe2015batch}}, to the entire mini-batch, can affect the generalization of large batch training. To ameliorate that, Ghost Batch Normalization \mbox{\cite{hoffer2017train}} was proposed, where a large batch is first scattered over multiple small virtual (``ghost'') batches. Then, each sample is normalized with respect to the mean and standard deviation of the ghost batch to which it belongs to. At inference, normalization is done via a weighted running average that aggregates the statistics of all ghost batches during training.

Similarly, concerned with the dependency of batch-normalized activations on the entire mini-batch, Salimans \textit{et. al} \mbox{\cite{salimans2016weight}} proposed Weight Normalization, which operates on the weight vectors instead of activations. Hence, it does not introduce any dependencies between the samples within a mini-batch. At core, it decouples the norm ($\mathbf{g}$) from the direction ($\frac{\mathbf{v}}{\lVert\mathbf{v}\rVert}$) of the weight vectors. Specifically, considering $\mathbf{w}$ as a weight vector in a standard artificial neural network, weigh normalization performs optimization over $\mathbf{g}$ and $\mathbf{v}$ using  $\mathbf{w}=\frac{\mathbf{g}}{\lVert\mathbf{v}\rVert}\mathbf{v}$ reparameterization. Salimans \textit{et. al} \mbox{\cite{salimans2016weight}} present weight normalization as a cheaper and less noisy approximation to the batch normalization \mbox{\cite{ioffe2015batch}} because first, convolutional neural networks usually have much fewer weights than activations and second, norm of $\mathbf{v}$ is non-stochatic but mini-batch statistics can have high variance for small batch sizes. Unfortunately, the guarantees that \mbox{\cite{salimans2016weight}} provides on the activations and gradients do not extend to models with arbitrary non-linearities or when the architecture contains layers without weight normalization \mbox{\cite{ioffe2017batch}}.}

\subsection{Batch Normalization: Extended}\label{sec:extended_normalization}
Recently, a few extensions on batch normalization have been proposed, specifically, Layer Normalization (LN) \cite{ba2016layer}, Instance Normalization (IN) \cite{ulyanov2017improved}, Group Normalization (GN) \cite{wu2018group}, and Divisive Normalization (DN) \cite{ren2016normalizing}. In this section, we loosely adopt notations from \cite{ren2016normalizing} and \cite{wu2018group} to show what distinguishes all the aforementioned methods is solely the set on which sample statistics are computed. This unifying view was initially presented by Ren \textit{et al.}\cite{ren2016normalizing} in an attempt to describe the relationship between BN, LN, and DN. Here we further extend it to IN and GN as well. 

Let's consider $i=(i_{N},i_{C},i_{L})$ as a vector indexing the tensor of activations $x\in\mathop{\mathbb{R}^{N\times C\times L}}$ associated to a convolution layer, where the spatial domain has been flattened ($L=H\times W$). Then the general normalization, $x\to\hat{x}$, is in form of
\begin{equation}\label{eq:general_form_normalization}
    v_{i} = x_{i}-\mathbb{E}_{\mathcal{B}_{i}}[x],\quad
    \hat{x}_{i} = \frac{v_{i}}{\sqrt{\mathbb{E}_{\mathcal{B}_{i}}[v^2]+\epsilon}},
\end{equation}
while similar to BN, $\gamma$ and $\beta$ parameters can be further applied to the normalized activations. Before describing the extensions to the batch normalization using Equation \ref{eq:general_form_normalization}, it is worth pointing out that given
\begin{equation}
    \mathcal{B}_{i}=\{j:j_{N}\in{[1,N]},j_{C}\in{[i_{C}]},j_{L}\in{[1,L]}\},
\end{equation}
the general normalization form reduces to the original batch normalization formulated in Equations \ref{eq:batch_normalization1} and \ref{eq:batch_normalization2}.

\textbf{Layer Normalization (LN)} was proposed by Ba \textit{et al.} \cite{ba2016layer}, where they remove the inter-dependency of batch-normalized activations to the entire mini-batch. Despite its effectiveness in recurrent networks, LN underperforms BN when applied to the convolution layers. That is because LN enforces the same distribution on the entire spatial domain \textit{and} along channel axis which is not natural in case of convolution layers as the visual information can dramatically vary over the spatial domain. LN can be formulated as Equation \ref{eq:general_form_normalization} when 
\begin{equation}
    \mathcal{B}_{i}=\{j:j_{N}\in{[i_{N}]},j_{C}\in{[1,C]},j_{L}\in{[1,L]}\}.
\end{equation}

\textbf{Instance Normalization (IN)} was proposed by Ulyanov \textit{et al.} \cite{ulyanov2017improved} for the problem of image style transfer \cite{gatys2016image}. While enjoying no inter-dependency to other samples in the mini-batch, IN \cite{ulyanov2017improved} uses more relaxed conditions than LN \cite{ba2016layer}, by computing the statistics only over the spatial domain generating different mean and standard deviations for each sample \textit{and} each channel. IN can be formulated as Equation \ref{eq:general_form_normalization} when 
\begin{equation}
    \mathcal{B}_{i}=\{j:j_{N}\in{[i_{N}]},j_{C}\in{[i_{C}]},j_{L}\in{[1,L]}\}.
\end{equation}

\textbf{Group Normalization (GN)} \cite{wu2018group} is somewhere in between LN and IN. GN divides the channels into multiple groups ($G=32$ by default), then computes the statistics along $L$ axis but only within a subgroup of the channels. Therefore, when the number of groups matches the channel size ($G=C$), GN is identical to IN. On the other hand, when there is only one group ($G=1$), GN reduces to LN. GN has shown to be effective on image classification, object detection and segmentation, when batch size is very small (2 and 4) while providing comparably good results with BN on typical batch sizes. GN can be formulated as Equation \ref{eq:general_form_normalization} when 
\begin{equation}
\begin{split}
    \mathcal{B}_{i}=\{j:j_{N}\in{[i_{N}]},j_{C}\in{[1,C]},j_{L}\in{[1,L]}\mid\\ \lfloor\frac{j_{C}}{C/G}\rfloor=\lfloor\frac{i_{C}}{C/G}\rfloor\},
\end{split}
\end{equation}
where $\lfloor\frac{j_{C}}{C/G}\rfloor$ ensures that normalizing $x_{i}$ is only influenced by the activation vectors which fall within the same group as $x_{i}$. 

\textbf{Divisive Normalization (DN)} \cite{ren2016normalizing} can be seen as a local version of LN, where in normalizing $x_{i}$, instead of all the activation vectors within the same layer, only those that are in a certain vicinity of $x_{i}$ contribute. This very well addresses the aforementioned drawback of LN, when applied to the convolution layer. DN is formulated slightly different from the aforementioned normalization methods, specifically as
\begin{equation}\label{eq:divisive_normalization}
    v_{i} = x_{i}-\mathbb{E}_{\mathcal{A}_{i}}[x]\quad
    \hat{x}_{i} = \frac{v_{i}}{\sqrt{\rho^2+\mathbb{E}_{\mathcal{B}_{i}}[v^2]}},
\end{equation}
when
\begin{equation}
    \mathcal{A}_{i}=\{j:d(x_{i},x_{j})\le R_{\mathcal{A}}\}\quad
    \mathcal{B}_{i}=\{j:d(v_{i},v_{j})\le R_{\mathcal{B}}\},
\end{equation}
where $d$ denotes an arbitrary distance between two hidden units, $\rho$ is the normalizer bias, and $R$ denotes the neighbourhood radius. Ren \textit{et al.} \cite{ren2016normalizing} have shown how varying $\rho$ would allow DN followed by ReLU to alternate within a wide range of non-linear behaviors. DN shows promising results both on convolutional and recurrent networks and can be easily implemented as a convolutional operator where $R_{\mathcal{A}}$ and $R_{\mathcal{B}}$ are determined by the kernel size.
\section{Proposed Method}\label{sec:method}
In this section, we show how the formulation of batch normalization and its extensions is related to the kernels from generative probability models, and more specifically Fisher kernel. Then, we explain why in the context of deep neural networks, due to the non-linearities, distribution of the activations is almost certainly of multiples modes of variation. Finally, we detail our proposed mixture normalization, where we employ Gaussian mixture model to identify sub-populations and normalize with respect to not one, but multiple components that comprise the data distribution.

\subsection{Kernels from Generative Probability Models}\label{sec:generative_probability_models}
So far, we've described variations of batch normalization in an unifying framework and shown what distinguishes them is the set of samples over which certain statistics, specifically mean and standard deviation are computed. In this section, we revisit Fisher kernel, the seminal work of Jaakkola and Haussler \cite{jaakkola1999exploiting}, in order to demonstrate how the general form of normalization, specified in Equation \ref{eq:general_form_normalization}, is related to the kernels that naturally arise from the generative probability models. 

\paragraph*{\textbf{Fisher Kernel}}
Let $\mathcal{B}=\{x_{1\dots m}\}$, regardless of how the tensor of activations has been indexed (ref. Section \ref{sec:extended_normalization}), be a set of $m$ observations $x_{i}\in\mathscr{X}$, associated to a sample mini-batch, and $\mathcal{P}=\{p_{\theta}, \theta\in\Theta\}$ be a suitably regular parametric family of distributions, where $p_{\theta}$ models the generative process of samples in $\mathscr{X}$\footnote{\hlcyan{In other words, $\mathscr{X}$ is an unknown generative process which is modeled by $p_{\theta}$, and we assume that instances in a mini-batch are sampled from it.}}. Based on the theory of information geometry \cite{amari2007methods}, $\mathcal{P}$ defines a Riemanninan manifold $M_{\Theta}$, with a local metric given by the Fisher Information Matrix
\begin{equation}\label{eq:fisher_information_matrix}
    F_{\theta} = \mathbb{E}_{x\sim p_{\theta}}[G_{\theta}^{\mathcal{B}}G_{\theta}^{\mathcal{B}^T}],
\end{equation}
where
\begin{equation}\label{eq:fisher_score}
    G_{\theta}^{\mathcal{B}} = \nabla_{\theta} \log p_{\theta}(\mathcal{B}),
\end{equation}
the gradient of the log-likelihood of $p_{\theta}$ at $\mathcal{B}$, also known as \textit{Fisher score}, determines the direction of steepest ascent in $\log p_{\theta}(\mathcal{B})$ along the manifold. In other words, $ G_{\theta}^{\mathcal{B}}$ describes modifications which the current model parameters $\theta$ need in order to maximize $\log p_{\theta}(\mathcal{B})$ \cite{jaakkola1999exploiting}. The natural gradient \cite{amari1998natural} is then defined as
\begin{equation}
    \phi_{\mathcal{B}}=F_{\theta}^{-1}G_{\theta}^{\mathcal{B}},
\end{equation}
based on which \cite{jaakkola1999exploiting} proposes natural kernel, the inner product between natural gradients relative to the local Riemannian metric: 
\begin{equation}\label{eq:fisher_kernel}
    K(\mathcal{B}_{j},\mathcal{B}_{i})\propto\phi_{\mathcal{B}_{j}}^T F_{\theta}\phi_{\mathcal{B}_{i}} = G_{\theta}^{\mathcal{B}_{j}^T}F_{\theta}^{-1}G_{\theta}^{\mathcal{B}_{i}}.
\end{equation}
Jaakkola and Haussler \cite{jaakkola1999exploiting} refer to this as \textit{Fisher kernel} and prove that given classification labels as latent variable, Fisher kernel is asymptotically never inferior to the maximum \textit{a posteriori} (MAP) decision rule (ref. Theorem 1 in \cite{jaakkola1999exploiting}). Hence, we arrive at a similarity measure which is naturally induced from the probability density function that models the generative process of $\mathscr{X}$ \textit{and} simultaneously improves the discrimination power of the model. 

Sanchez \textit{et. al} \cite{sanchez2013image} have beautifully pointed out that since $F_{\theta}$ is positive semi-definite, its inverse has a Cholesky decomposition $F_{\theta}^{-1} = L_{\theta}^T L_{\theta}$, which allows re-writing Equation \ref{eq:fisher_kernel} as
\begin{equation}\label{eq:fisher_vector0}
K(\mathcal{B}_{j},\mathcal{B}_{i}) = \mathscr{G}_{\theta}^{\mathcal{B}_{j}^T}\mathscr{G}_{\theta}^{\mathcal{B}_{i}},
\end{equation}
where
\begin{equation}\label{eq:fisher_vector}
    \mathscr{G}_{\theta}^{\mathcal{B}} = L_{\theta}\nabla_{\theta} \log p_{\theta}(\mathcal{B}),
\end{equation}
the normalized gradient vector of $\mathcal{B}$, is known as \textit{Fisher vector} \cite{sanchez2013image}. Such explicit transformation $\mathcal{B}\to\mathscr{G}_{\theta}^{\mathcal{B}}$ enjoys all the interesting characteristics of Fisher kernel \cite{jaakkola1999exploiting}, while being tailored for linear operations. 

\paragraph*{\textbf{Fisher Vector for Gaussian Distribution}}
In order to derive Fisher kernel \cite{jaakkola1999exploiting}, we noted that $p_{\theta}$ only needs to be of a family of regular parametric distributions.
However, we have not yet parameterized it. Let's consider $p_{\theta}$ to be modelled by Gaussian distribution 
\begin{equation}
        p_{\theta}(x) = \frac{1}{(2\pi)^{D/2}\lvert \Sigma \rvert^{1/2}}\,\text{exp}\bigg\{-\frac{1}{2}{(x-\mu)}^T \Sigma^{-1}(x-\mu)\bigg\},
\end{equation}
where $x\in\mathbb{R}^D$ and $\theta=\{\mu,\Sigma\}$ is the set of model parameters. The gradients of the log-likelihood of $p_{\theta}(x)$ with respect to $\mu$ and $\Sigma$ are then formulated as
\begin{equation}
    \begin{matrix*}
    \nabla_{\mu} \log p_{\theta}(x) &= -\frac{1}{2}\Big(\frac{\partial {(x-\mu)}^T \Sigma^{-1}(x-\mu)}{\partial \mu}\Big) = \Sigma^{-1}(x-\mu),\\
    \nabla_{\Sigma} \log p_{\theta}(x) &=
    -\frac{1}{2}\Big(\frac{\partial \log(\lvert \Sigma \rvert)}{\partial \Sigma}+\frac{\partial {(x-\mu)}^T \Sigma^{-1}(x-\mu)}{\partial \Sigma}\Big),\\
    &=-\frac{1}{2}\Big(\Sigma^{-1}-\Sigma^{-1}(x-\mu){(x-\mu)}^T\Sigma^{-1}\Big).
    \end{matrix*}
\end{equation}
If we assume $\Sigma$ to be diagonal \footnote{\hlcyan{Such assumption is motivated by the fact that computing full covariance matrix and its inverse makes both the normalization and the back-propagation calculations very expensive \mbox{\cite{ioffe2015batch}}.}} with $\Sigma_{i,i}=\sigma_{i,i}^2$, we can re-write the gradients as
\begin{equation}\label{eq:partial_derivatives}
    \begin{split}
    \nabla_{\mu} \log p_{\theta}(x) &= \frac{1}{\sigma^2}(x-\mu),\\
    \nabla_{\sigma} \log p_{\theta}(x) &=
    -\frac{1}{\sigma}+\frac{1}{\sigma^3}{(x-\mu)}^2.
    \end{split}
\end{equation}
Substituting Equation \ref{eq:partial_derivatives} in Equation \ref{eq:fisher_score} results in 
\begin{equation}\label{eq:fisher_score_gaussian}
    G_{\theta}^{x} = \begin{bmatrix} 
    \nabla_{\mu} \log p_{\theta}(x) \\ \nabla_{\sigma} \log p_{\theta}(x) \end{bmatrix}=\begin{bmatrix}
    \dfrac{1}{\sigma^2}(x-\mu)\\  -\dfrac{1}{\sigma}+\dfrac{1}{\sigma^3}{(x-\mu)}^2\end{bmatrix}.
\end{equation}
We then use non-central moments of univariate normal distribution to calculate Equation \ref{eq:fisher_information_matrix} given Equation \ref{eq:fisher_score_gaussian}, arriving at
\begin{equation}\label{eq:fisher_information_matrix_gaussian}
    F_{\theta} = \begin{bmatrix} \dfrac{1}{\sigma^2} & 0 \\ 0 &  \dfrac{2}{\sigma^2}\end{bmatrix},
\end{equation}
the Fisher information matrix of Gaussian distribution. Finally, the Fisher vector for Gaussian distribution (ref. Equation \ref{eq:fisher_vector}) can be characterized as
\begin{equation}\label{eq:fisher_vector_gaussian}
\begin{split}
    \mathscr{G}_{\theta}^{x} &=
    \begin{bmatrix}\mathscr{G}_{\mu}^{x}\\  \mathscr{G}_{\sigma}^{x}\end{bmatrix},\\
    &=\begin{bmatrix} \sigma & 0 \\ 0 & \frac{1}{\sqrt{2}}\sigma\end{bmatrix}\begin{bmatrix}\dfrac{1}{\sigma^2}(x-\mu)\\-\dfrac{1}{\sigma}+\dfrac{1}{\sigma^3}{(x-\mu)}^2\end{bmatrix},\\
    &=\begin{bmatrix}\dfrac{1}{\sigma}(x-\mu)\\  \dfrac{1}{\sqrt{2}}\Big(-1 + \dfrac{(x-\mu)^2}{\sigma^2}\Big)\end{bmatrix}.
\end{split}
\end{equation}

We observe that the general form of normalization, formulated in Equation \ref{eq:general_form_normalization}, has in fact emerged in $\mathscr{G}_{\mu}^{x}$. Hence, we have shown that indeed batch normalization and its various extensions are closely related to the natural kernel that arises from generative probability model, describing the underlying data distribution \footnote{\hlcyan{Based on our derivations, we ``expect'' that integrating $\mathscr{G}_{\sigma}^{x}$ in addition to $\mathscr{G}_{\mu}^{x}$ further improves the normalization. Yet, at the time, we do not have experimental results to support this.}}. 

\subsection{Mixture Normalization}\label{sec:mixture_normalization}
We believe that in the context of deep neural networks, due to non-linear activation functions, distribution of layer outputs are very likely asymmetric. That is, the hypothesis which a Gaussian distribution can model $p_{\theta}$ is less likely to be valid. Therefore, to properly approximate the probability density function, we propose employing generative \textit{mixture} models. Consequently, instead of computing one set of statistical measures from the entire population, we propose to frame batch normalizing transform on sub-populations which can be identified by disentangling modes of variation.

\paragraph*{\textbf{Intuition}}Let's consider $x_{l}\in\mathop{\mathbb{R}^{N\times C_{l}\times H_{l}\times W_{l}}}$ to be the input activation tensor to the $l^{th}$ layer, denoted by $\omega_{l}$, in a convolutional neural network. Again, $N$, $C_{l}$, $H_{l}$ and $W_{l}$ are respectively batch, channel, height and width axes at the corresponding layer. Batch normalization layers are indexed similarly. Since the batch size is fixed throughout the network, we drop the subscript $l$ from $N_{l}$. For the sake of simplicity, we ignore pooling layers and assume the non-linearity activation functions to be rectified linear units (ReLU) \cite{nair2010rectified}, since pooling layers can be trivially added to the formulation. Similarly, other non-linearity functions can partially or fully replace ReLU throughout the network. Given these,

\begin{equation}\label{eq:three_consecutive}
\begin{matrix*}[l]
    x_{l-1}&=&\mathop{\text{ReLU}\Big(\text{BN}_{l-2}(x_{l-2}\ast \omega_{l-2})\Big)},\\
    x_{l}&=&\mathop{\text{ReLU}\Big(\text{BN}_{l-1}(x_{l-1}\ast \omega_{l-1})\Big)},\\
    x_{l+1}&=&\mathop{\text{ReLU}\Big(\text{BN}_{l}(x_{l}\ast \omega_{l})\Big)},
\end{matrix*}    
\end{equation}
formulate three consecutive layers in the aforementioned convolution neural network where $*$ represents the convolution operation. In Section \ref{sec:generative_probability_models}, given the hypothesis $x\sim\mathcal{N}(\mu,\,\sigma^{2})$, the general form of normalization in Equation \ref{eq:general_form_normalization} emerges as the Fisher vector with respect to the mean of the distribution, namely $\mathscr{G}_{\mu}^{x}$. This means that the input to $\text{BN}_{l}$ at the $l^{th}$ layer in Equation \ref{eq:three_consecutive}, specifically $x_{l}\ast \omega_{l}$, is of a Gaussian distribution. Since, convolution is a linear operation, and Gaussian distribution is closed under linear combination, it further implies that $x_{l}\sim\mathcal{N}(\mu_{l},\,\sigma_{l}^{2})$. However, $x_{l}$ is the output of ReLU from the $(l-1)^{th}$ layer with semi-infinite support on $[0,+\infty)$. Hence, assuming that a symmetric probability density function, like Gaussian distribution, with support on the whole Real line can model the linear combination of outputs of multiple rectified linear units, is not very well justified. 

Based on this observation, we propose to parameterize the probability density function, $p_{\theta}$ (ref. Section \ref{sec:generative_probability_models}), as a Gaussian Mixture Model (GMM). From \cite{titterington1985statistical}, we know that any continuous distribution can be approximated with arbitrary precision using a GMM. Since our input distribution is generated by linear combination (through convolution operation) of rectified Gaussian distributions, it is not necessarily continuous and contains, at least two modes, one for the rectified values mapped to zero and one for the positive values. Therefore, we expect a mixture model to provide us with a better approximation of such distribution than a single Gaussian model.

\paragraph*{\textbf{Fisher Vector for Gaussian Mixture Model}}In the following, to prevent clutter, we drop subscript $\theta$ where $\theta=\{\lambda_{k},\mu_{k},\Sigma_{k}:k=1,\dots,K\}$. Then, for $x\in\mathbb{R}^D$, without loss of generality, consider
\begin{equation}\label{eq:gmm0}
    p(x) = \mathop{\mathbb{\sum}}_{k=1}^{K}\lambda_{k} p_{k}(x)\quad
    s.t.\,\forall_{k}: \lambda_{k}\ge 0,\,\mathop{\mathbb{\sum}}_{k=1}^{K}\lambda_{k}=1,
\end{equation}
where 
\begin{equation}
        p_{k}(x) = \frac{1}{(2\pi)^{D/2}\lvert \Sigma_{k} \rvert^{1/2}}\,\text{exp}\bigg\{-\frac{1}{2}{(x-\mu_{k})}^T \Sigma_{k}^{-1}(x-\mu_{k})\bigg\},
\end{equation}
represents $k^{th}$ Gaussian in the mixture model $p(x)$. To remove the probability simplex constraint on $\lambda$ in Equation \ref{eq:gmm0}, a commonly used change of variables \cite{jordan1994hierarchical} is employed as
\begin{equation}
    \alpha_{k} = \log\Bigg(\frac{\lambda_{k}}{\lambda_{K}}\Bigg),\, k=1,\dots,K.
\end{equation}
Assume $\alpha_{K}=0$ to be constant, then we can re-write Equation \ref{eq:gmm0} as 
\begin{equation}\label{eq:gmm}
    p(x) = \mathop{\mathbb{\sum}}_{k=1}^{K}\Big(\frac{\text{exp}(\alpha_{k})}{\mathop{\mathbb{\sum}}_{j=1}^{K}\text{exp}(\alpha_{j})}\Big)\,p_{k}(x).
\end{equation}
Following similar derivations as in Section \ref{sec:generative_probability_models}, given diagonal covariance assumption, the gradients of the log-likelihood of $p(x)$ with respect to $\mu_{k}$ and $\sigma_{k}$ can be written as
\begin{equation}\label{eq:partial_derivatives_gmm}
    \begin{split}
    &\nabla_{\mu_{k}} \log p(x) = \nu_{k}(x)\Bigg(\dfrac{x-\mu_{k}}{\sigma_{k}^2}\Bigg),\\
    &\nabla_{\sigma_{k}} \log p_{\theta}(x) =
    \nu_{k}(x)\Bigg(-\dfrac{1}{\sigma_{k}}+\dfrac{(x-\mu_{k})^2}{\sigma_{k}^3}\Bigg).
    \end{split}
\end{equation}
where
\begin{equation}\label{eq:posterior}
    \nu_{k}(x) = \frac{\lambda_{k} p_{k}(x)}{\mathop{\mathbb{\sum}}_{j=1}^{K}\lambda_{j} p_{j}(x)},
\end{equation}
is the probability that $x$ has been generated by the $k^{th}$ Gaussian component in the mixture model. Given soft assignment distribution in Equation \ref{eq:posterior}, Perronnin \textit{et. al} \cite{perronnin2007fisher} proposed a closed form approximation to the Fisher information matrix of GMM\footnote{For more detailed derivation of Fisher information matrix and Fisher vector in Gaussian mixture models, readers are encouraged to refer to \cite{perronnin2007fisher} and \cite{sanchez2013image}.} as
\begin{equation}\label{eq:fisher_information_matrix_gmm}
    F_{k} = \begin{bmatrix} \dfrac{\lambda_{k}}{\sigma_{k}^2} & 0 \\ 0 & \dfrac{2\lambda_{k}}{\sigma_{k}^2}\end{bmatrix},
\end{equation}
resulting in 
\begin{equation}\label{eq:fisher_vector_gmm}
\begin{split}
    \mathscr{G}_{k}^{x} &=
    \begin{bmatrix}\mathscr{G}_{\mu_{k}}^{x}\\  \mathscr{G}_{\sigma_{k}}^{x}\end{bmatrix},\\
    &=\begin{bmatrix} \dfrac{\sigma_{k}}{\sqrt{\lambda_{k}}} & 0 \\ 0 & \dfrac{\sigma_{k}}{\sqrt{2\lambda_{k}}}\end{bmatrix}
    \begin{bmatrix}\nu_{k}(x)\Bigg(\dfrac{x-\mu_{k}}{\sigma_{k}^2}\Bigg)\\\nu_{k}(x)\Bigg(-\dfrac{1}{\sigma_{k}}+\dfrac{(x-\mu_{k})^2}{\sigma_{k}^3}\Bigg)\end{bmatrix},\\
    &=\dfrac{\nu_{k}(x)}{\sqrt{\lambda_{k}}}\begin{bmatrix}\dfrac{1}{\sigma_{k}}(x-\mu_{k})\\\dfrac{1}{\sqrt{2}}\Bigg(-1+\dfrac{(x-\mu_{k})^2}{\sigma_{k}^2}\Bigg)\end{bmatrix}.
\end{split}
\end{equation}

Comparing Equation \ref{eq:fisher_vector_gmm} with Equation \ref{eq:fisher_vector_gaussian}, we observe that Fisher vector for each component of a GMM is very similar to the one obtained from a single Gaussian distribution. However, there are two main differences:
\begin{itemize}
    \item $\nu_{k}(x)$ in a soft-assignment mechanism, scales the $k^{th}$ Fisher vector based on the posterior probability of $x$ being generated by the corresponding Gaussian component.
    \item $\lambda_{k}$ normalizes the $k^{th}$ Fisher vector based on the contribution of the $k^{th}$ mixture component in approximating the underlying data distribution.
\end{itemize}

We have previously shown (ref. Section \ref{sec:generative_probability_models}) that the normalization formulated in batch normalization and its extensions emerges as natural kernel, \textit{if} we assume that the underlying data distribution can be modelled by a single Gaussian distribution. We then explained that due to non-linear activation functions, such hypothesis cannot properly address the characteristics of the distribution. We employed Gaussian mixture model as an alternative and showed that natural kernel induced from GMM, follows a similar normalization mechanism, except normalization is done with respect to multiple sets of statistics obtained from different sub-populations.

\paragraph*{\textbf{Formulation}}Based on the aforementioned observations, we propose Mixture Normalization. Let's consider $i=(i_{N},i_{C},i_{L})$ as a vector indexing the tensor of activations $x\in\mathop{\mathbb{R}^{N\times C\times L}}$ associated to a convolution layer, where the spatial domain has been flattened ($L=H\times W$). Then the \textit{Mixture Normalizing Transform} is defined as
\begin{equation}\label{eq:mixture_normalization0}
    \hat{x}_{i} = \mathop{\mathbb{\sum}}_{k=1}^{K}\dfrac{\nu_{k}(x_{i})}{\sqrt{\lambda_{k}}}\cdot\hat{x}_{i}^{k},
\end{equation}
given
\begin{equation}\label{eq:mixture_normalization1}
    v_{i}^{k} = x_{i}-\mathbb{E}_{\mathcal{B}_{i}}[\hat{\nu}_{k}(x)\cdot x],\quad
    \hat{x}_{i}^{k} = \frac{v_{i}^{k}}{\sqrt{\mathbb{E}_{\mathcal{B}_{i}}[\hat{\nu}_{k}(x)\cdot {(v^k)}^2]+\epsilon}},
\end{equation}
where
\begin{equation}\label{eq:responsibility}
    \hat{\nu}_{k}(x_{i}) = \frac{\nu_{k}(x_{i})}{\mathop{\mathbb{\sum}}_{j\in\mathcal{B}_{i}}\nu_{k}(x_{j})},
\end{equation}
is the normalized contribution of $x_{i}$ over $\mathcal{B}_{i}$ in estimating statistical measures of the $k^{th}$ Gaussian component. Similar to batch normalization, we can also include additional parameters by slightly modifying Equation \ref{eq:mixture_normalization1} using
\begin{equation}
    \begin{matrix*}[l]
    \mathbb{E}_{\mathcal{B}_{i}}[\hat{\nu}_{k}(x)\cdot x]&\to& \mathbb{E}_{\mathcal{B}_{i}}[\hat{\nu}_{k}(x)\cdot x] + \beta,\\
    \mathbb{E}_{\mathcal{B}_{i}}[\hat{\nu}_{k}(x)\cdot {(v^k)}^2]&\to& \mathbb{E}_{\mathcal{B}_{i}}[\hat{\nu}_{k}(x)\cdot {(v^k)}^2] \cdot \gamma,
    \end{matrix*}
\end{equation}
where $\beta$ and $\gamma$ respectively indicate scale and shift. Our formulation is general (\textit{e.g.}  $H,W=1$ for fully-connected layers) and can be applied to all the variations of batch normalization, detailed in Section \ref{sec:extended_normalization}, simply by constructing $\mathcal{B}_{i}$ accordingly.

\hlcyan{During training, we fit a Gaussian mixture model to $\mathcal{B}=\{x_{1\dots m}:m\in{[1,N]\times[1,L]}\}$ by Maximum Likelihood Estimation (MLE). This is a two-stage process where we use the seeding procedure of K-means++ \mbox{\cite{arthur2007k}} to initialize the centers of the mixture component. Then, the parameters of the mixture model, $\theta=\{\lambda_{k},\mu_{k},\Sigma_{k}:k=1,\dots,K\}$, are estimated by Expectation-Maximization (EM) \mbox{\cite{dempster1977maximum}}. In practice, one or two EM iterations are sufficient, thanks to proper and efficient initialization by K-means++ \mbox{\cite{arthur2007k}}. We then normalize $x_1, x_2, \dots x_m$ with respect to the estimated parameters (Equations \mbox{\ref{eq:mixture_normalization1}}) and aggregate using posterior probabilities (Equations \mbox{\ref{eq:mixture_normalization0}}). At inference, following the batch normalization \mbox{\cite{ioffe2015batch}}, we want to normalize a sample mini-batch with respect to the statistics of the training data. To do so, we can maintain last $T$ mini-batches from the training stage. This can be simply implemented by a queue of length $T$. We aggregate all instances in the queue into a large pool of samples and fit a Gaussian mixture model to it. Estimated parameters are then used to normalize all the mini-batches of the test set. While being effective, the above strategy is not very desirable as it always maintains $T$ mini-batch of samples in GPU memory. Yet, there is a very simple workaround to significantly improve this. Specifically, during training, instead of the samples, we only need to keep the estimated parameters resulting in the queue $\{\theta_{0},\theta_{1}\dots \theta_{T-1}\}$, where the subscript indicates the position in the queue. Note that in this case, queue always maintains $(2KC+K)\times T$ values in the GPU memory which is orders of magnitude smaller than $(mC) \times T$ of the previous strategy. The normalization will then be performed via

\begin{equation}\label{eq:mixture_normalization_test}
    \hat{x}_{i} = \mathop{\mathbb{\sum}}_{t=0}^{T-1}\mathop{\mathbb{\sum}}_{k=1}^{K} \dfrac{1}{\sqrt{\tau^{t}\lambda^{t}_{k}}} \frac{\tau^{t}\lambda^{t}_{k} p^{t}_{k}(x_{i})}{\mathop{\mathbb{\sum}}_{q=0}^{T-1}\mathop{\mathbb{\sum}}_{j=1}^{K} \tau^{q}\lambda^{q}_{j} p^{q}_{j}(x_{i})}\Big(\dfrac{x_{i}-\mu^{t}_{k}}{\sigma^{t}_{k}}\Big),
\end{equation}
where
\begin{equation}\label{eq:decay_factor}
\tau^{t} = \frac{(1-\zeta)}{(1-\zeta^T)} \zeta^{(T-t-1)}
\end{equation}
is the decay factor normalized with respect to the sum of its geometric series. Here, estimated mixture models from last $T$ mini-batch of training stage are aggregated with a scale proportional to their chronological order in the queue. By default, $\zeta$ and $T$ are respectively set to 0.9 and 10.}

% During training, we always maintain recent mini-batches in a fixed length (20 as default) queue. At inference, we sample elements of the queue with a rate proportional to their chronological order. Specifically, sampling rate (0.9 as default) decays exponentially as we move from the most recent to earlier mini-batches. Then, we fit a GMM model on the aggregated population once, and use the mixture model to normalize all the mini-batches at inference. The queue will be emptied at the beginning of each epoch. Estimating the GMM parameters is done via Expectation Maximization (EM) \cite{dempster1977maximum}, where we use K-means clustering to initialize the solution. 

\paragraph*{\textbf{\hlcyan{Differentiability and Gradient Propagation}}} \hlcyan{Despite our end-to-end training, K-means++ \mbox{\cite{arthur2007k}} seeding procedure and EM iterations (to fit the Gaussian mixture model) are performed outside the computational graph of the neural network\footnote{In future, we are going to explore mixture normalization, when estimating GMM via batch and stochastic Riemannian optimization \mbox{\cite{hosseini2017alternative}}.}. Hence, MN is not fully differentiable. In the following, we explain how gradient back-propagates through MN.

Let $\mathbf{\nu}\in\mathbb{R}^{m\times K}$ be a matrix where its $(i,j)^{th}$ element represents the probability that $x_i$, the $i^{th}$ row of $\mathbf{x}\in\mathbb{R}^{m\times D}$ ($\mathcal{B}$ in matrix format), has been generated by the $j^{th}$ Gaussian component in the mixture model (Equation \mbox{\ref{eq:posterior}}). The process of obtaining $\mathbf{\nu}$ is not differentialble because of K-means++ and following EM iterations, yet we can write MN equations such that with exception of $\mathbf{\nu}$, it is fully differentiable.

Following Equation \mbox{\ref{eq:responsibility}}, we obtain the normalized contribution of each sample point $x_i$ in estimating statistical measures of every Gaussian component via dividing each column of $\mathbf{\nu}$ by the sum of its elements. This results in $\mathbf{\hat{\nu}}$, the matrix of responsibilities using which we calculate $\mathbf{\mu}=\mathbf{\hat{\nu}}^{T}\mathbf{x}$ and $\mathbf{\sigma}^{2}=\mathbf{\hat{\nu}}^{T}\mathbf{x}^{2}-\mathbf{\mu}^{2}$ where $\mathbf{\mu}\in \mathbb{R}^{K\times D}$, $\mathbf{\sigma}^{2}\in \mathbb{R}^{K\times D}$, and $\lambda_{k}$ is obtained by averaging the $k^{th}$ column of $\mathbf{\nu}$. As we can see, all these calculations are simple differentiable operations. Hence, while gradient stops at $\mathbf{\nu}$, it seamlessly back-propagates through the rest of MN equations, relating parameters of the mixture component with samples and their posterior.}

\paragraph*{\textbf{Effect of Non-linearity}} When mixture normalization is followed by ReLU \cite{nair2010rectified} or similar non-linearity activation functions, rectifying should be applied to the per-component normalized activations (\textit{i.e.} $\hat{x}_{i}^{k}$) as
\begin{equation}\label{eq:relu_mixturenorm}
    \mathop{\text{ReLU}\Big(\text{MN}(x_{i})\Big)}\coloneqq\mathop{\mathbb{\sum}}_{k=1}^{K}\dfrac{\nu_{k}(x_{i})}{\sqrt{\lambda_{k}}}\cdot\text{ReLU}(\hat{x}_{i}^{k}),
\end{equation}
where MN stands for mixture normalization. In practice, given $K\ge3$,  $\nu(x_{i})=\big[\nu_1(x_{i}),\,\nu_2(x_{i}),\dots,\, \nu_K(x_{i})\big]$ is mostly very sparse, meaning that only 1 out of $K$ elements is considerably larger than zero. As a result, we have
\begin{equation}\label{eq:relu_mixturenorm_appr}
    \mathop{\mathbb{\sum}}_{k=1}^{K}\dfrac{\nu_{k}(x_{i})}{\sqrt{\lambda_{k}}}\cdot\text{ReLU}(\hat{x}_{i}^{k})\approx\mathop{\text{ReLU}\Big(\mathop{\mathbb{\sum}}_{k=1}^{K}\dfrac{\nu_{k}(x_{i})}{\sqrt{\lambda_{k}}}\cdot\hat{x}_{i}^{k}\Big)},
\end{equation}
a format which is the same as how ReLU follows batch normalization. However, when Gaussian components considerably overlap, the aforementioned approximation is less accurate. 
% We have also observed experimentally, that the approximate formulation introduces some small instabilities.

\begin{figure*}[htbp]
    \centering
    \begin{subfigure}[b]{\textwidth}
        \includegraphics[width=\textwidth]{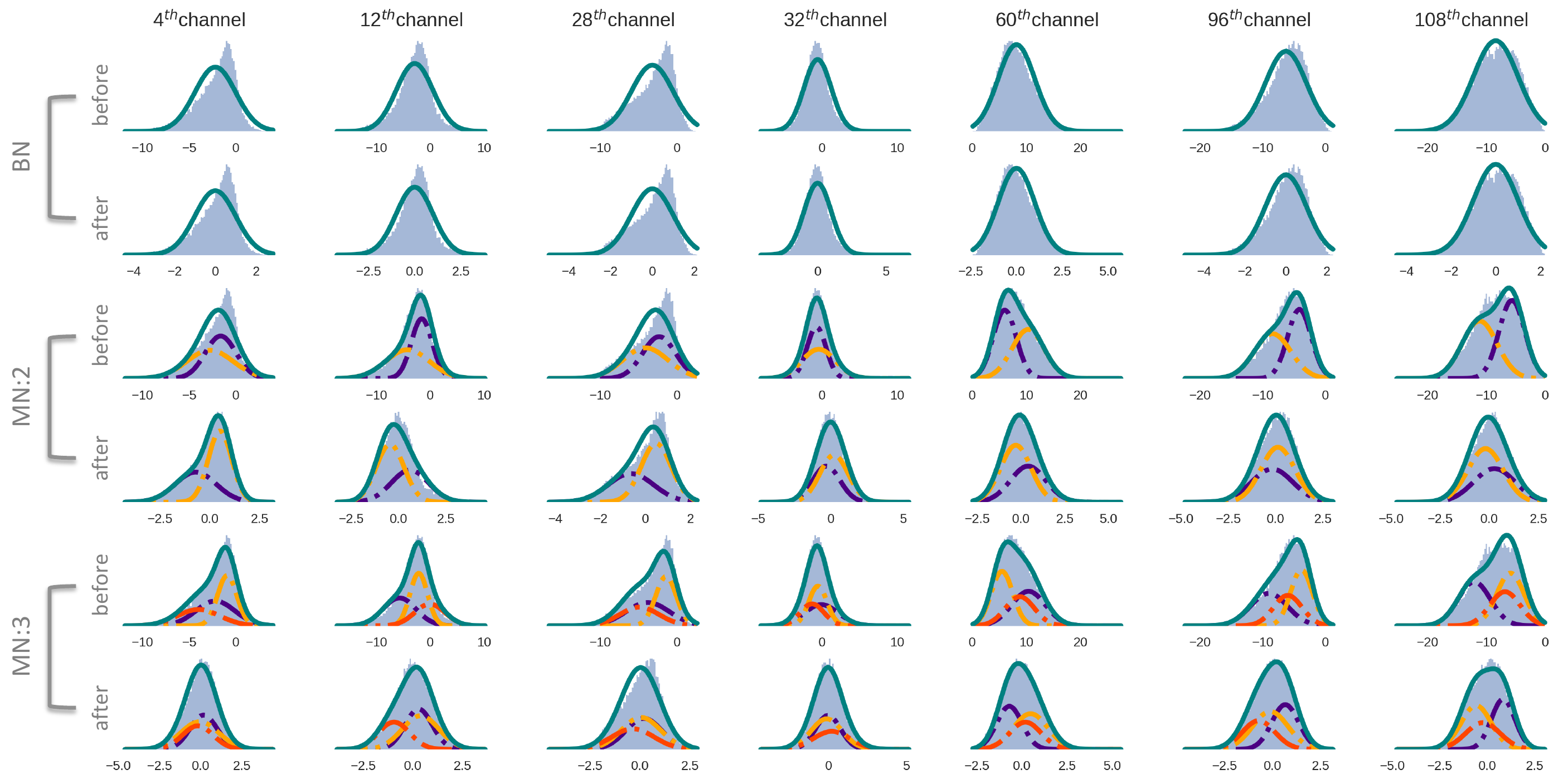}
    \end{subfigure}
    \caption{Visualizing Mixture Normalization: Given a random mini-batch \hlcyan{in the midway of training} on CIFAR-100, we illustrate the underlying distribution of activations (output of convolution) associated with a random subset of 128 channels in the layer ``conv2'' of CIFAR CNN architecture (detailed in Table \ref{tab:cifar_cnn_arch}). Solid teal curve indicates the probability density function. Dashed curves represent different mixture components shown in various colors. Note that similar colors across multiple subfigures, simply index mixture components and do not indicate any association. We observe that mixture normalization, shown by MN:2 and MN:3 (2 and 3 respectively represent the number of components in the mixture of Gaussians), provides better approximation, $p(x)$, illustrated by solid teal curve, to the underlying distribution. Also, mixture normalized activations, in comparison to the batch normalized ones, are considerably closer to the normal distribution and illustrate less skewed probability densities.}\label{fig:teaser}
\end{figure*}

\paragraph*{\textbf{Local vs. Global Normalization}} Batch normalization is an Affine transform on the \textit{whole} probability density function of the underlying distribution. In other words, \textit{all} the samples from the distribution are normalized using the \textit{same} mean and standard deviation, estimated from the entire population. In contrast, in proposed mixture normalization, we transform each sample using mean and standard deviation of the mixture component to which the sample belongs to. Therefore, one can see mixture normalization as a soft piecewise normalizing transform. Figure \ref{fig:teaser} illustrates the aforementioned differences. Note how in case of batch normalization, $p(x)$, modeled by a Gaussian density function, fails to properly approximate the underlying data distribution. This problem is more severe when the distribution is skewed. Mixture normalization instead, is capable of handling these challenges as it disentangles the modes of variation. In Figure \ref{fig:teaser}, we show mixture normalization with $K$ set to 2 and 3. Obviously, when the number of components is set to 1, mixture normalization reduces to the general form of normalization. We observe that even with two components, mixture normalization provides a better approximation to the underlying data distribution. However, in some cases (\textit{e.g.} columns correspond to the $4^{th}$ and $28^{th}$ channels), we can benefit from increasing the number of mixture components. Mixture normalized activations, in comparison to the batch normalized ones, are considerably closer to the normal distribution, and illustrate less skewed probability densities.

\section{Experiments}\label{sec:experiments}
To evaluate the effectiveness of our proposed mixture normalization, we conduct extensive set of experiments on CIFAR-10 and CIFAR-100 datasets \cite{krizhevsky2009learning}. We compare mixture normalization against its batch normalization counterpart in a variety of settings given four backbone choices, namely a 5-layers deep fully convolutional neural network, the Inception-V3 \cite{szegedy2016rethinking}, along with the 40-layers and 100-layers deep DenseNet \cite{huang2017densely} architectures. According to the literature \cite{wu2018group}\cite{ren2016normalizing}\cite{ioffe2017batch}, when mini-batch size is sufficiently large (\textit{e.g} 32), standard batch normalization \cite{ioffe2015batch} outperforms its variants (ref. Section \ref{sec:extended_normalization}) in image classification. Although, this is not valid in recurrent networks. Therefore, here, we only compare against standard batch normalization \cite{ioffe2015batch}.

It is important to emphasize, that we do not aim at achieving state-of-the-art results as it requires employing computationally expensive architectures, and involves careful tuning of many hyper-parameters and heuristics. Instead, we are interested in understanding the behavior of mixture normalization. We focus on demonstrating that solely replacing \textit{a few} batch normalization with our proposed mixture normalization, can dramatically increase the convergence rate, and in majority of cases even yields a better final test accuracy. 

In summary, in this section:

\begin{itemize}
    \item We compare BN and MN, in small and large learning rate regimes, using both shallow and very deep CNN architectures.
    \item We show the effect of applying MN to different layers, while varying the number of mixture components and EM iterations.
    \item We quantify the acceleration obtained using MN, by measuring required number of gradient updates for mixture normalized model, in order to reach the best test accuracy of its batch normalized counterpart.
    \item We demonstrate, that our findings are consistent with respect to the choice of the optimization technique (adaptive v.s non-adaptive) and learning rate decay policies.
    \item We finish this section by demonstrating the application of mixture normalization in Generative adversarial networks (GANs) \cite{goodfellow2014generative}.
\end{itemize}

\subsection{Datasets}\label{sec:datasets}
For our experiments, we use popular CIFAR \cite{krizhevsky2009learning} datasets. Images are 32$\times$32 and labeled with 10 and 100 classes, respectively for CIFAR-10 and CIFAR-100. We follow the standard data split where 50K images are used for training and 10K for testing. As for preprocessing, we normalize images with respect to the mean and standard deviation of the dataset. Also, similar to the previous works \cite{he2016deep}\cite{huang2017densely}\cite{zagoruyko2016wide}\cite{huang2016deep}\cite{lin2013network} on CIFAR, we adopt horizontal flipping and random cropping for data augmentation.

\subsection{CIFAR CNN}\label{sec:cifar_cnn}
We begin with a small 5-layers deep convolutional neural network architecture detailed in Table \ref{tab:cifar_cnn_arch}, where we later replace batch normalization in certain layers with mixture normalization. Following \cite{ioffe2015batch}, we experiment using two different learning rates, one 5 times larger than the other. Additionally, we do the same with the weight decay to very well cover various training regimes. We use RMSprop \cite{tieleman2012lecture} with 0.9 momentum and exponentially reduce the initial learning rate every two epochs with the decay rate of 0.93. The size of mini-batch is set to 256 and we train all the models for 100 epochs. To facilitate comparison between different training settings, we index experiments accordingly, where BN and MN, respectively, indicate usage of batch and mixture normalization.

\begin{table}[htbp]
  \centering
  \begin{tabular}{cllcc}
	\toprule
    layer & type & size & kernel & (stride, pad)\\
    \midrule
    input & input & $3\times32\times32$ & -- & --\\
    conv1 & conv+bn+relu & $64\times32\times32$ & $5\times5$ & (1, 2)\\
    pool1 & max pool & $64\times16\times16$ & $3\times3$ & (2, 0)\\
    conv2 & conv+bn+relu & $128\times16\times16$ & $5\times5$ & (1, 2)\\
    pool2 & max pool & $128\times8\times8$ & $3\times3$ & (2, 0)\\
    conv3 & conv+bn+relu & $128\times8\times8$ & $5\times5$ & (1, 2)\\
    pool3 & max pool & $128\times4\times4$ & $3\times3$ & (2, 0)\\   
    conv4 & conv+bn+relu & $256\times4\times4$ & $3\times3$ & (1, 1)\\
    pool4 & avg pool & $256\times1\times1$ & $4\times4$ & (1, 0)\\   
    linear & linear & 10 or 100 & -- & -- \\   
    \bottomrule
  \end{tabular}
  \caption{CIFAR CNN architecture}
  \label{tab:cifar_cnn_arch}
\end{table}

\begin{table*}[htbp]
  \centering
  \begin{tabular}{l*{6}{c}}
	\toprule
	\multicolumn{7}{c}{\textbf{CIFAR-10}}\\
	\midrule  
    & \multicolumn{1}{c}{mixture norm. setting} & \multicolumn{1}{c}{training setting} & \multicolumn{4}{c}{maximum test accuracy(\%) after}\\
    \cmidrule(lr){2-2}\cmidrule(lr){3-3}\cmidrule(lr){4-7}
    model & (layer, K, EM iter.) & (lr, weight decay) & 25 epochs & 50 epochs & 75 epochs & 100 epochs\\
    \midrule
    BN-1 & -- & (0.01, 1e-4) & 79.37 & 84.34 & 86.24 & 86.74\\
    BN-2 & -- & (0.01, 2e-5) & 84.11 & 86.92 & 87.65 & 87.95\\
    BN-3 & -- & (0.05, 1e-4) & 68.77 & 71.68 & 75.25 & 77.35\\
    BN-4 & -- & (0.05, 2e-5) & 73.84 & 77.84 & 81.43 & 82.58\\
    \midrule
    MN-1 & (conv3, 3, 2) & (0.01, 1e-4) & 81.91 & 85.10 & 86.51 & 87.08\\
    MN-2 & (conv3, 3, 2) & (0.01, 2e-5) & 85.56 & 87.28 & 88.05 & 88.47\\
    MN-3 & (conv3, 3, 2) & (0.05, 1e-4) & 67.84 & 73.09 & 74.80 & 77.33\\
    MN-4 & (conv3, 3, 2) & (0.05, 2e-5) & 76.70 & 80.82 & 83.31 & 83.87\\
	\toprule
	\multicolumn{7}{c}{\textbf{CIFAR-100}}\\
	\midrule  
    & \multicolumn{1}{c}{mixture norm. setting} & \multicolumn{1}{c}{training setting} & \multicolumn{4}{c}{maximum test accuracy(\%) after}\\
    \cmidrule(lr){2-2}\cmidrule(lr){3-3}\cmidrule(lr){4-7}
    model & (layer, K, EM iter.) & (lr, weight decay) & 25 epochs & 50 epochs & 75 epochs & 100 epochs\\
    \midrule
    BN-1 & -- & (0.01, 1e-4) & 52.54 & 57.83 & 60.37 & 62.11\\
    BN-2 & -- & (0.01, 2e-5) & 56.09 & 60.05 & 61.46 & 62.14\\
    BN-3 & -- & (0.05, 1e-4) & 23.13 & 28.12 & 33.21 & 36.80\\
    BN-4 & -- & (0.05, 2e-5) & 36.98 & 43.89 & 46.95 & 49.08\\
    \midrule
    MN-1 & (conv3, 3, 2) & (0.01, 1e-4) & 55.30 & 60.47 & 61.46 & 62.20\\
    MN-2 & (conv3, 3, 2) & (0.01, 2e-5) & 59.04 & 60.86 & 61.77 & 62.29\\
    MN-3 & (conv3, 3, 2) & (0.05, 1e-4) & 29.80 & 36.30 & 40.66 & 43.77\\
    MN-4 & (conv3, 3, 2) & (0.05, 2e-5) & 42.31 & 48.96 & 51.98 & 52.62\\
    MN-5 & (conv3, 3, 4) & (0.01, 2e-5) & 59.05 & 61.94 & 62.56 & 62.76\\
    MN-6 & (conv3, 3, 8) & (0.01, 2e-5) & 58.98 & 61.83 & 62.67 & 63.12\\
    MN-7 & (conv2, 3, 2) & (0.01, 2e-5) & 57.63 & 61.23 & 62.15 & 62.63\\
    MN-8 & ((conv2,conv3), 3, 2) & (0.01, 2e-5) & 59.79 & 61.67 & 62.50 & 62.96\\
    \bottomrule
  \end{tabular}
  \caption{Experiments on CIFAR-10 and CIFAR-100 using CIFAR CNN architecture (ref. Table \ref{tab:cifar_cnn_arch}). We observe that irrespective of the weight decay and learning rate, not only MN models converge faster but also achieve better final test accuracy, compared to their corresponding BN counterparts. When mixture normalization is applied to multiple layers (\textit{i.e} MN-8), we use the same K and EM iter. values for all the corresponding layers.}
  \label{tab:cifar_cnn_results}
\end{table*}

\begin{figure*}[h]
    \centering
    \begin{subfigure}[b]{0.495\textwidth}
        \centering
        \includegraphics[width=0.495\linewidth]{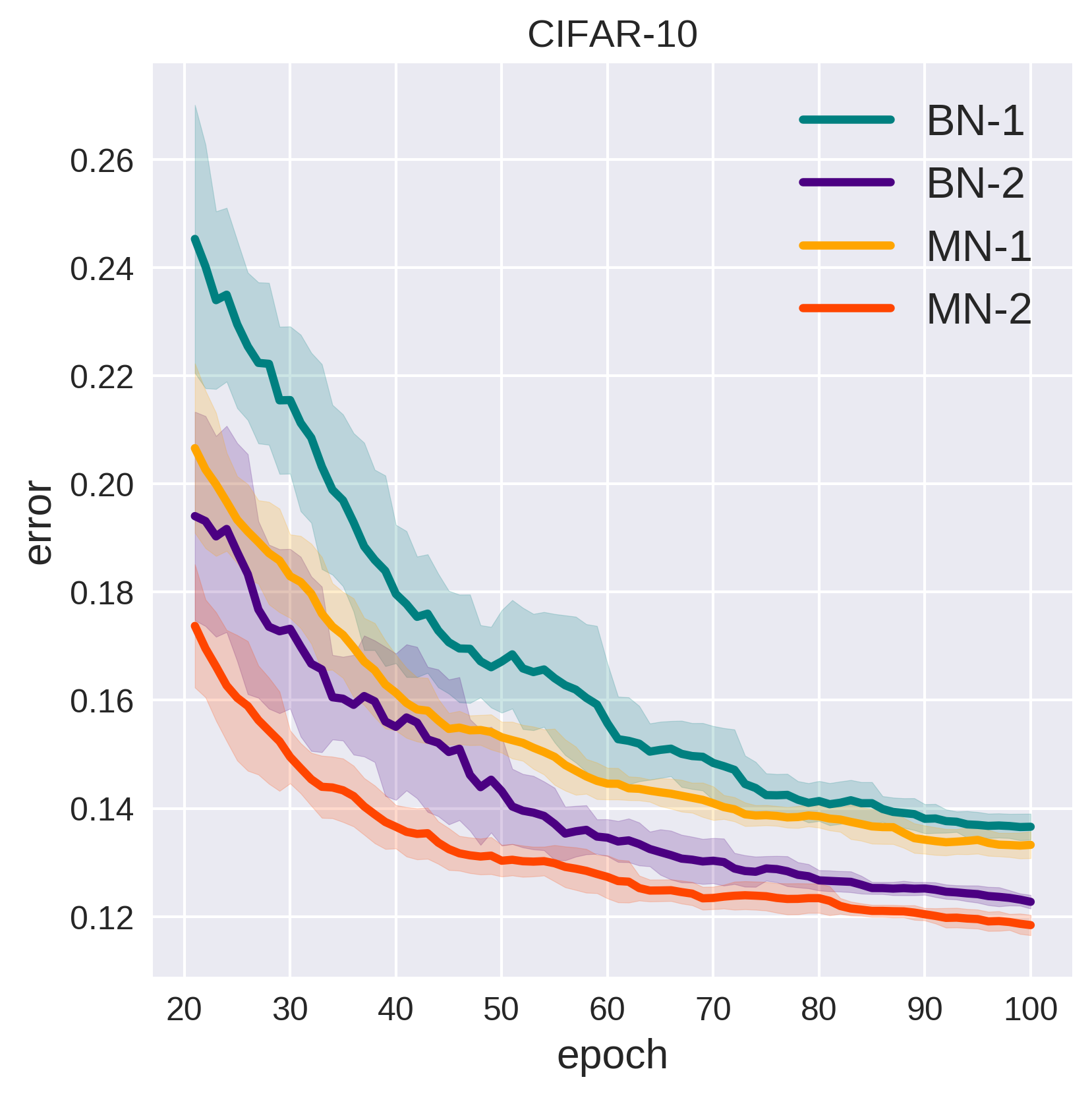}%
        \hfill
        \includegraphics[width=0.495\linewidth]{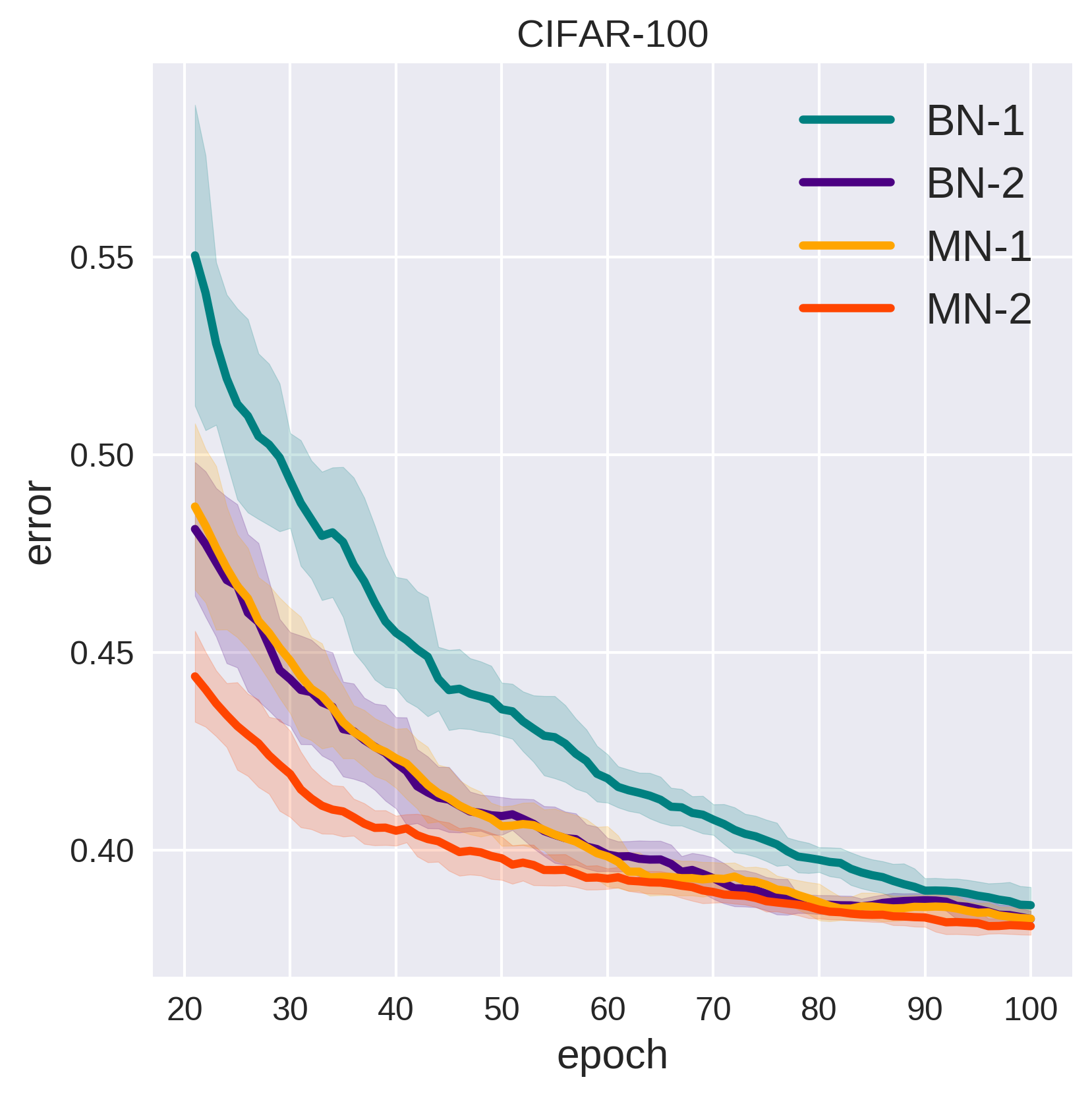}
        \caption{small learning rate regime}
        \label{fig:cifar_low_lr}
    \end{subfigure}
    \begin{subfigure}[b]{0.495\textwidth}
        \centering
        \includegraphics[width=0.495\linewidth]{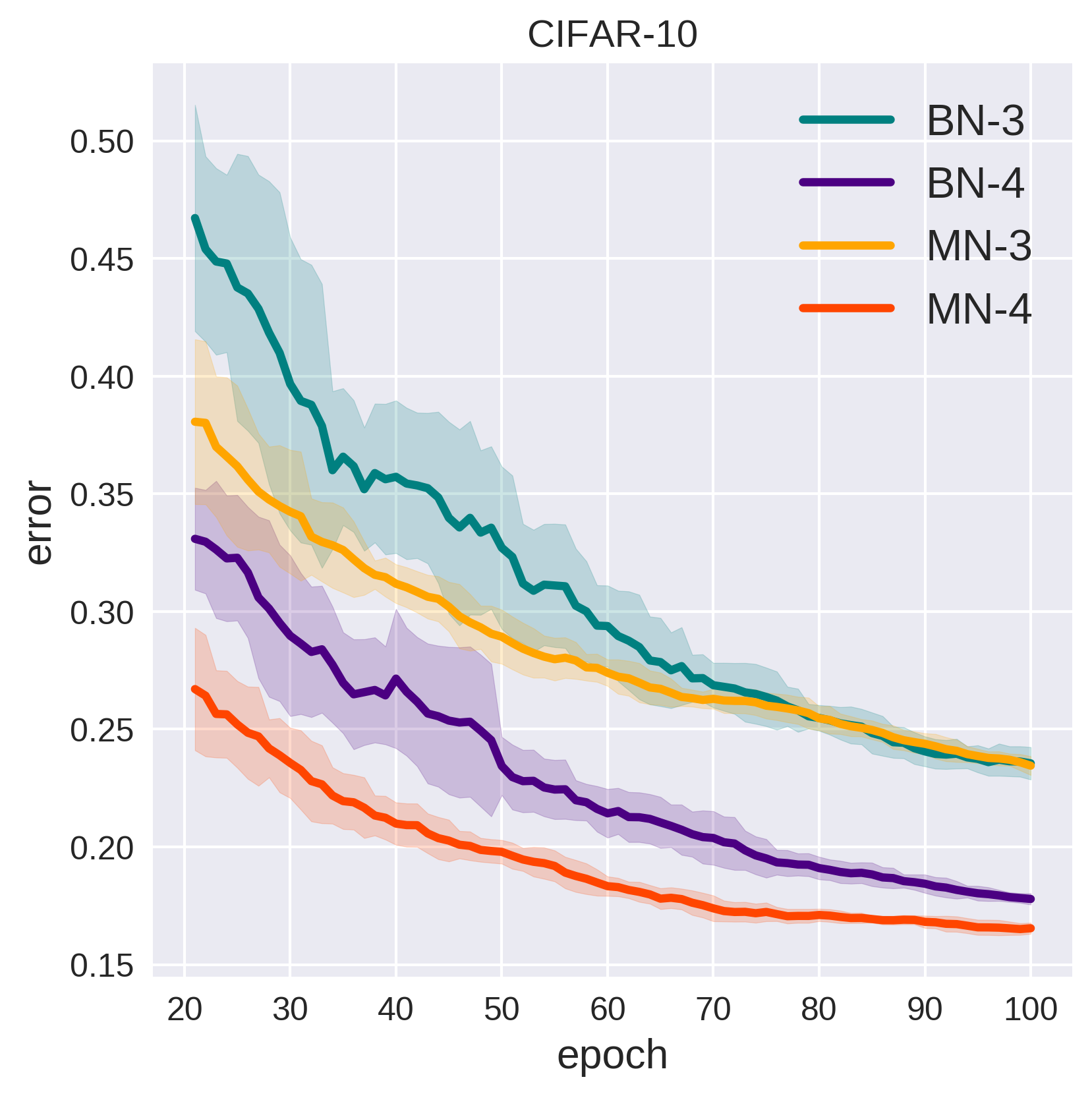}%
        \hfill
        \includegraphics[width=0.495\linewidth]{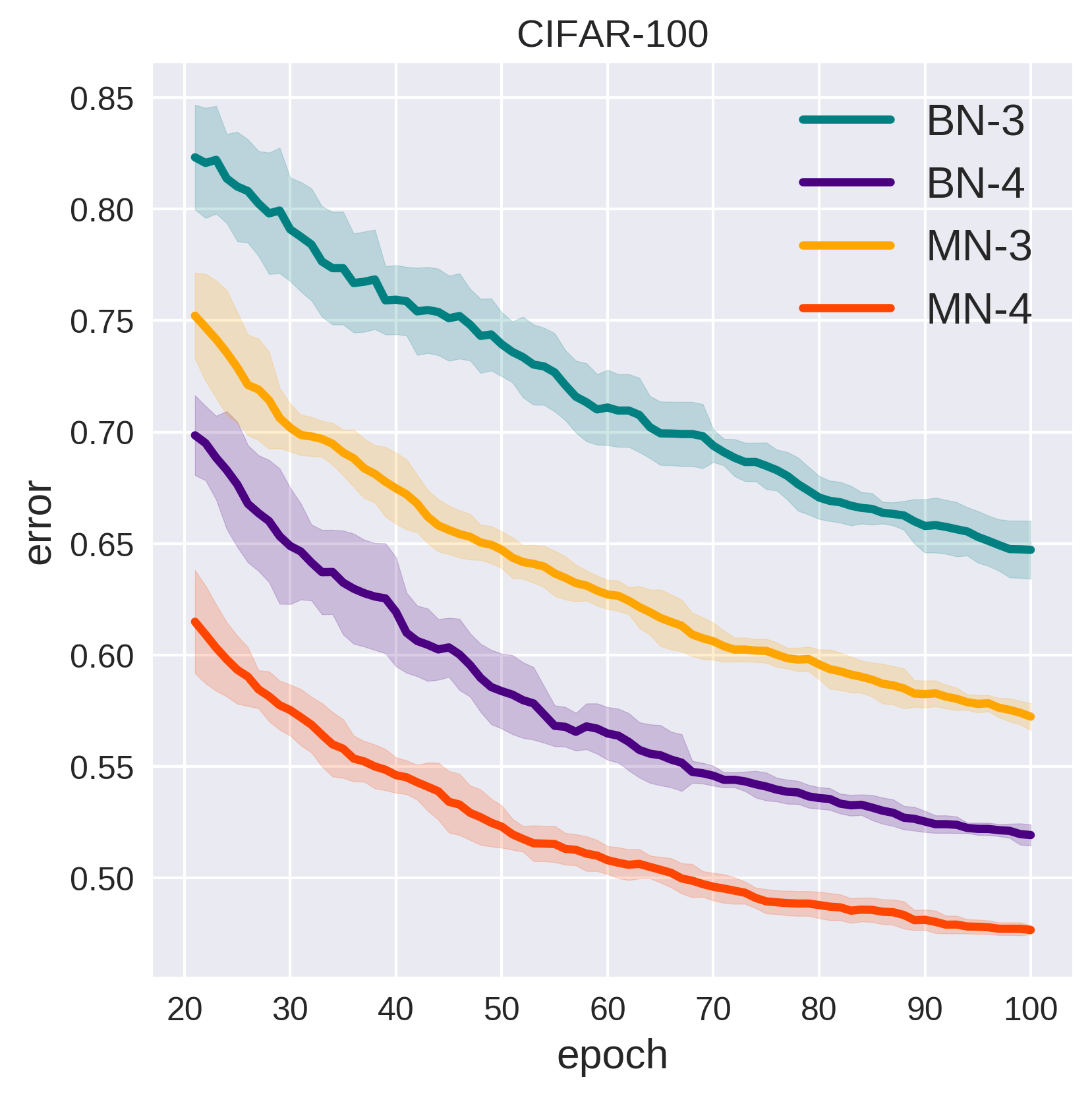}
        \caption{large learning rate regime}
        \label{fig:cifar_high_lr}
    \end{subfigure}
    \caption{Test error curves when CIFAR CNN architecture (ref. Table \ref{tab:cifar_cnn_arch}) is trained under different learning rate and weight decay settings. We observe that on CIFAR-10 and CIFAR-100, MN performs consistently in both small and large learning rate regimes.}\label{fig:cifar_cnn_results}
\end{figure*}

In the first set of experiments, for different mixture normalization variations, shown in Table \ref{tab:cifar_cnn_results} by MN-1 to MN-4, we replace the batch normalization in the ``conv3'' layer (ref. Table \ref{tab:cifar_cnn_arch}) with mixture normalization while keeping the remaining layers intact. From Table \ref{tab:cifar_cnn_results} and Figure \ref{fig:cifar_cnn_results}, we observe that irrespective of the weight decay and learning rates, not only MN models converge much faster than their corresponding BN counterparts, but they achieve a better test accuracy after 100 epochs. 

\textbf{Large learning rate:} When we increase the learning rate from 0.01 to 0.05, the convergence gap is even larger (ref. Table \ref{tab:cifar_cnn_results} and Figure \ref{fig:cifar_cnn_results}), demonstrating the capability of mixture normalization to better utilize larger learning rates for training. Note, that for the large learning rate regime to eventually match the final learning rate of small learning rate regime, either the exponential decay rate should be reduced (\textit{e.g} 0.91 instead of 0.93) or the number of times it is applied must be increased (\textit{e.g} every epoch instead of every 2 epochs). However, here we keep the same learning rate decay policy to probe the sole effect of increasing the learning rate. 

\textbf{Stability:} Another observation is with regards to the uncertainty of the model predictions. In Figure \ref{fig:cifar_cnn_results}, we show one standard deviation (shaded area) computed within a window of 10 epochs for all the test error curves. We can see that models trained with mixture normalization illustrate a considerably more stable test error, meaning that, from one epoch to another, it is less likely that the performance fluctuates abruptly. This alongside faster convergence \cite{hardt2016train} is reminiscent of relatively flat minima in the optimization landscape \cite{arora2018stronger}, where the classification margin of the model, as a whole, is robust with respect to small changes (gradient updates) to the model parameters. Note that in all these experiments, we have solely replaced one batch normalization layer with mixture normalization. 

\begin{figure}[htbp]
\begin{subfigure}{.495\textwidth}
  \centering
  \includegraphics[width=0.495\linewidth]{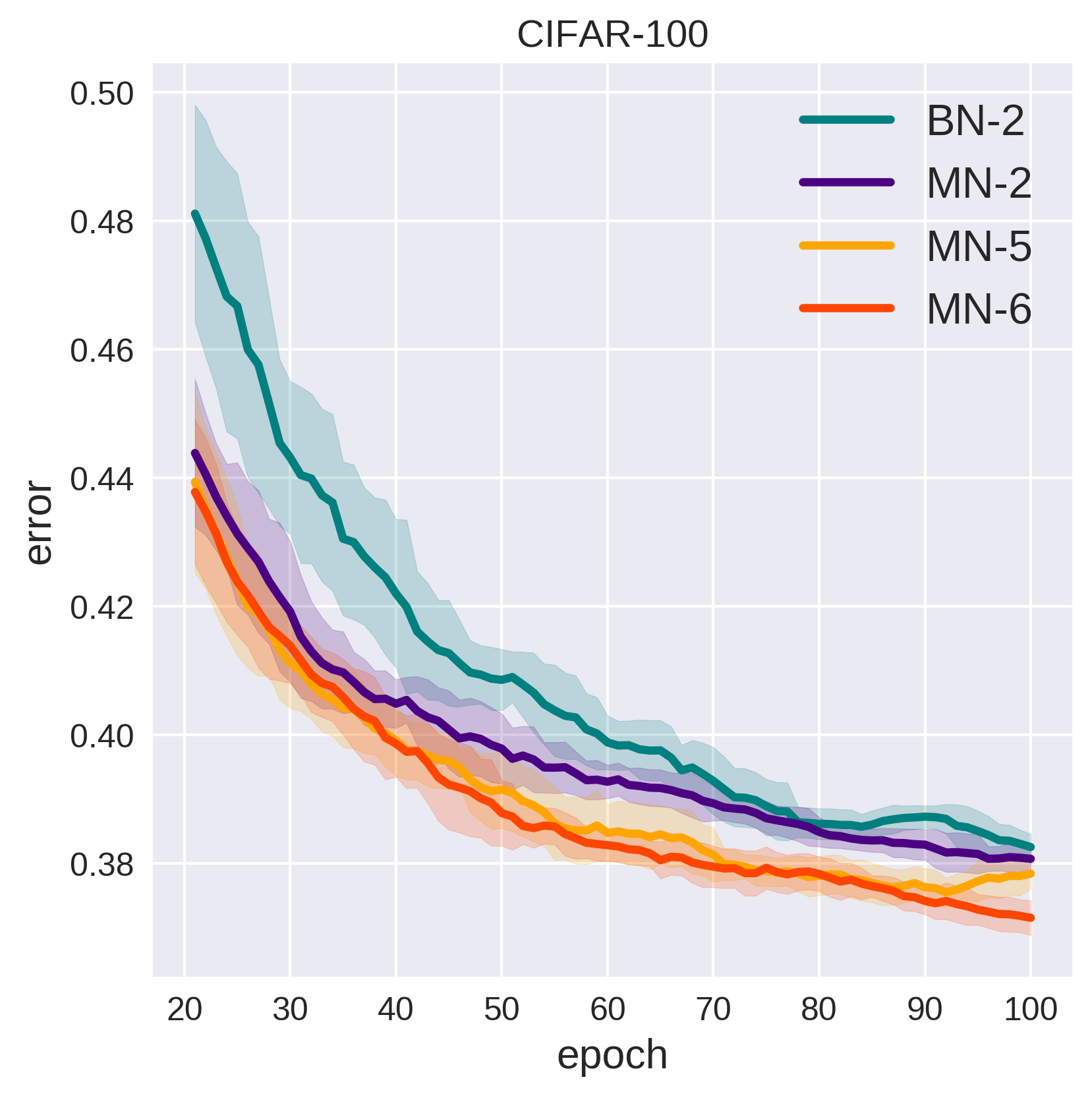}
  \hfill
  \includegraphics[width=0.495\linewidth]{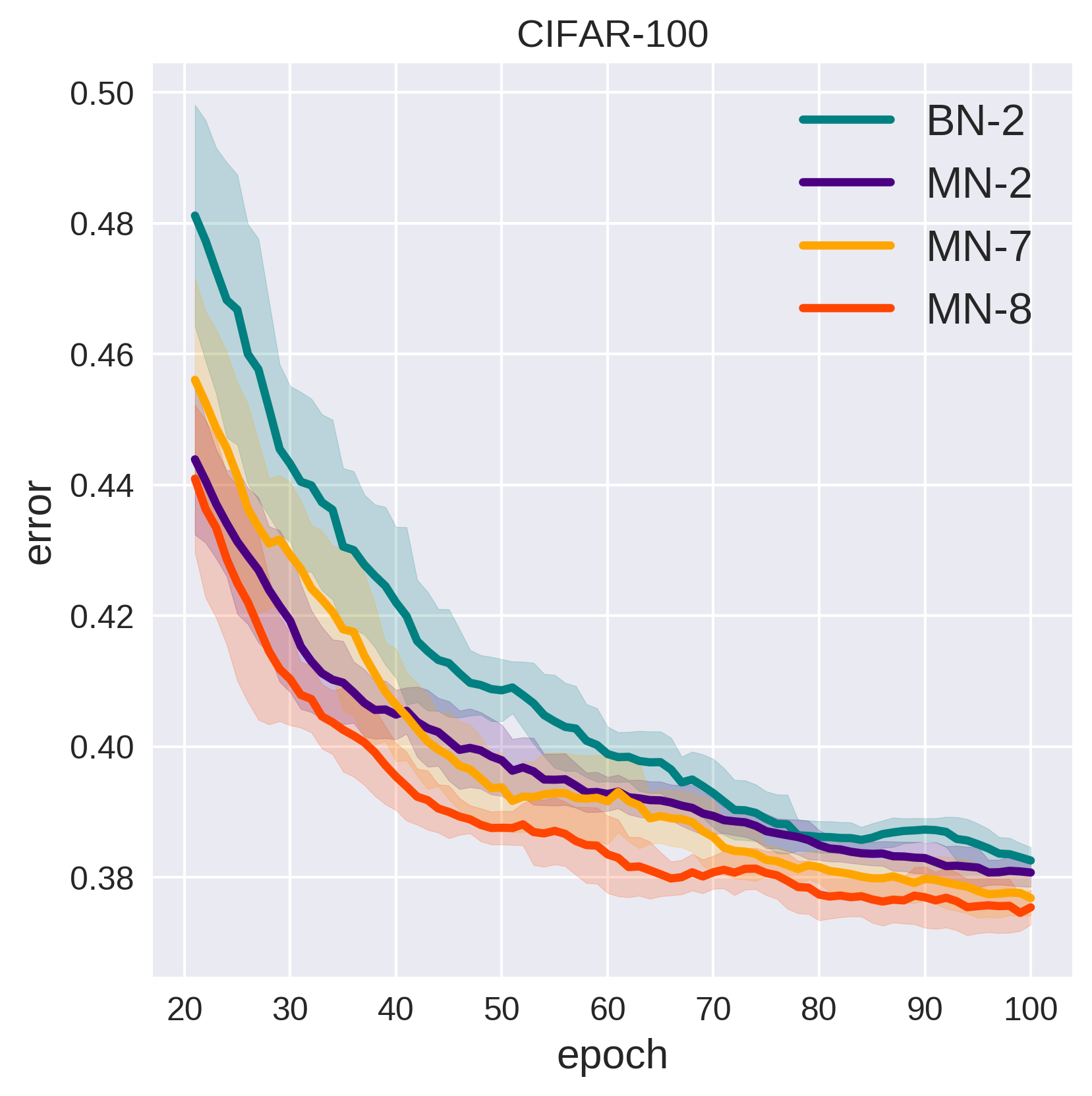}
\end{subfigure}
\caption{Left: effect of the number of EM iterations on test error. Right: effect of utilizing MN at different layers, on test error. We show that more EM iterations and utilizing MN at multiple layers, increase the convergence rate of mixture normalized models.}
\label{fig:cifar_cnn_mixture_params}
\end{figure}

\textbf{Effect of parameters:} Solving for the parameters of GMM begins with an initialization using K-means clustering. To obtain the best initial seeds, we use $2+\log(K)$ trials and then perform standard K-means clustering for a certain number of iterations. The result is then used to initiate the GMM parameters, which is later followed by a fixed number of EM updates. In our experiments, we use the same number of iterations at two phases and report their summation as ``EM iter.'' in Table \ref{tab:cifar_cnn_results} (and later in Table \ref{tab:cifar_inceptionv3_results}). From Figure \ref{fig:cifar_cnn_mixture_params} (left) and Table \ref{tab:cifar_cnn_results}, we observe that more EM iterations, provides faster convergence and better final test accuracy. This is aligned with our expectation as more EM iterations translates to better approximation of the underlying data distribution. However, the downside is that more EM iterations results in increasing the computation time. So far, we have only modified one layer in CIFAR CNN architecture. In Figure \ref{fig:cifar_cnn_mixture_params} (right), we illustrate that similar behavior can be observed when mixture normalization is employed at earlier layers like ``conv2''. Furthermore, model enjoys even faster convergence rate in addition to better final test accuracy (ref. Table \ref{tab:cifar_cnn_results}), when more layers (``conv2'' and ``conv3'') are equipped with mixture normalization.

\begin{table}[htbp]
  \centering
  \begin{tabular}{l*{2}{c}}
	\toprule
	\multicolumn{3}{c}{\textbf{CIFAR-10}}\\
	\midrule  
    model & steps to 87.95\% & max. acc.(\%)\\
    \midrule
    BN-2 & 1.95 $\times$ $10^4$ & 87.95\\
    MN-2 & 1.34 $\times$ $10^4$ & 88.47\\
	\toprule
	\multicolumn{3}{c}{\textbf{CIFAR-100}}\\
	\midrule  
    model & steps to 62.14\% & max. acc.(\%)\\
    \midrule
    BN-2 & 1.81 $\times$ $10^4$ & 62.14\\
    MN-2 & 1.77 $\times$ $10^4$ & 62.29\\
    MN-5 & 1.07 $\times$ $10^4$ & 62.76\\
    MN-6 & 1.09 $\times$ $10^4$ & 63.12\\
    MN-7 & 1.34 $\times$ $10^4$ & 62.63\\
    MN-8 & 1.03 $\times$ $10^4$ & 62.96\\
    \bottomrule
  \end{tabular}
  \caption{For batch normalization and the mixture-normalized
variants using CIFAR CNN architecture (ref. Table \ref{tab:cifar_cnn_arch}), the number of training steps required to
reach the maximum accuracy of batch-normalized model alongside with the maximum accuracy achieved by each variant.}
  \label{tab:cifar_cnn_stepsto}
\end{table}

\textbf{Quantifying Acceleration:} We illustrate in Figure \ref{fig:cifar_cnn_results}, that indeed using mixture normalization accelerates the convergence and results in lower final test error. Table \ref{tab:cifar_cnn_results} also indicates the same when we report the maximum test accuracy after 25\%, 50\%, 75\% and 100\% of total number of epochs. To provide a precise comparison, we follow Ioffe and Szegedy \cite{ioffe2015batch} and report the number of steps (gradient descent updates) for mixture normalization variants in order to reach the maximum test accuracy achieved by their batch-normalized counterparts. Table \ref{tab:cifar_cnn_stepsto} shows that on CIFAR-10, mixture normalization reduces the number of training steps in order to reach 87.95\% test accuracy by $\sim$\textbf{31}\%. Similarly, on CIFAR-100, the best performing variant of mixture normalization, MN-6, reduces the number of training steps in order to reach 62.14\% test accuracy by $\sim$\textbf{40}\%, meanwhile its very own maximum test accuracy outperforms BN-2 by $\sim$1\%. This further affirms that, our proposed mixture normalization is not only effectively accelerating the training procedure, but also reaches better local minima.

\subsection{Inception-V3}\label{sec:cifar_inceptionv3}
So far, we have shown that mixture normalization can improve the convergence rate of batch-normalized models. However, our experiments were conducted on a shallow 5-layers deep architecture. Hence, it is reasonable to question whether the same behavior can be observed in very deep and more modern architectures? To address this, we choose Inception-V3 \cite{szegedy2016rethinking}. Its architecture is 48 layers deep and uses global average pooling instead of fully-connected layers, which allows operating on arbitrary input image sizes. Inception-V3 \cite{szegedy2016rethinking} has a total output stride of 32. However, to maintain low computation cost and memory utilization, the size of activation maps quickly reduces by a factor of 8 in only first seven layers. This is done by one convolution and two max pooling layers that operate with the stride of 2. The network is followed by three blocks of Inception separated by two grid reduction modules. Each Inception block consists of multiple Inception layers that are sequentially stacked. Specifically, first, second and third Inception blocks are respectively comprised of 3, 4 and 2 Inception layers. Spatial resolution of the activations remains intact within the Inception blocks, while grid reduction modules halve the activation size and increase the number of channels. 

To make Inception-V3 \cite{szegedy2016rethinking} architecture effectively applicable to images in CIFAR-10 and CIFAR-100, that are only 32$\times$32, we need to slightly modify the architecture. Specifically, we change the stride of the first convolution layer from 2 to 1 and remove the first max pooling layer. This way, the output stride of Inception-V3 architecture reduces to 8. That is, activations maintain sufficient resolution throughout the network's depth, with the final activation (before the global average pooling) be of size 3$\times$3. From now on, when we refer to the Inception-V3, we mean this modified version. 

To train our models, we use RMSprop \cite{tieleman2012lecture} with 0.9 as momentum and exponentially reduce the initial learning rate every four epochs with the decay rate of 0.93. The size of mini-batch is set to 128, weight decay to 0.0005 and we train all the models for 200 epochs. In our preliminary experiments, we observed that learning rate of 0.001 gives the best final test accuracy for batch normalized models. Therefore, we use it for all the experiments except in one case which aims at analyzing large learning rate regime.

\begin{table*}[htbp]
  \centering
  \begin{tabular}{l*{6}{c}}
	\toprule
	\multicolumn{7}{c}{\textbf{CIFAR-10}}\\
	\midrule  
    & \multicolumn{1}{c}{mixture norm. setting} & \multicolumn{1}{c}{training setting} & \multicolumn{4}{c}{maximum test accuracy(\%) after}\\
    \cmidrule(lr){2-2}\cmidrule(lr){3-3}\cmidrule(lr){4-7}
    model & (layer, K, EM iter.) & (lr, weight decay) & 50 epochs & 100 epochs & 150 epochs & 200 epochs\\
    \midrule
    BN-1 & -- & (0.001, 5e-4) & 89.18 & 90.62 & 91.08 & 91.50\\
    \midrule
    MN-1 & ((red1,red2), 3, 8) & (0.001, 5e-4) & 89.47 & 91.03 & 91.70 & 91.91\\
    MN-2 & ((red1,red2,inc3/0), 3, 4) & (0.001, 5e-4) & 89.73 & 91.69 & 92.09 & 92.55\\
    MN-3 & ((inc2/0,inc3/0), 3, 2) & (0.001, 5e-4) & 90.49 & 91.52 & 92.17 & 92.30\\
    MN-4 & ((inc2/0,inc3/0), 5, 2) & (0.001, 5e-4) & 90.26 & 91.77 & 92.00 & 92.25\\
	\toprule
	\multicolumn{7}{c}{\textbf{CIFAR-100}}\\
	\midrule  
    & \multicolumn{1}{c}{mixture norm. setting} & \multicolumn{1}{c}{training setting} & \multicolumn{4}{c}{maximum test accuracy(\%) after}\\
    \cmidrule(lr){2-2}\cmidrule(lr){3-3}\cmidrule(lr){4-7}
    model & (layer, K, EM iter.) & (lr, weight decay) & 50 epochs & 100 epochs & 150 epochs & 200 epochs\\
    \midrule
    BN-1 & -- & (0.001, 5e-4) & 66.88 & 69.44 & 71.03 & 71.30\\
    BN-5 & -- & (0.005, 5e-4) & 40.68 & 47.47 & 51.13 & 52.63\\
    \midrule
    MN-1 & ((red1,red2), 3, 8) & (0.001, 5e-4) & 66.6 & 68.86 & 70.01 & 70.43\\
    MN-2 & ((red1,red2,inc3/0), 3, 4) & (0.001, 5e-4) & 68.26 & 70.78 & 71.78 & 72.23\\
    MN-4 & ((inc2/0,inc3/0), 5, 2) & (0.001, 5e-4) & 68.4 & 70.99 & 71.98 & 72.74\\
    MN-5 & ((inc2/0,inc3/0), 5, 2) & (0.005, 5e-4) & 50.41 & 55.69 & 57.88 & 59.10\\
    \bottomrule
  \end{tabular}
  \caption{Experiments on CIFAR-10 and CIFAR-100 using Inception-V3 architecture. Notation: ``red1'' (``red2'') refers to the first (second) grid reduction modules. Similarly ``inc2/0'' (``inc3/0'') refers to the first inception layer in second (third) inception block of the architecture. In MN-$\ast$ settings, we only replace the last batch normalization in each branch of the corresponding Inception layer with our mixture normalization. When mixture normalization is applied to multiple layers, we use the same K and EM iter. values for all the corresponding layers.}
  \label{tab:cifar_inceptionv3_results}
\end{table*}

\begin{figure*}[h]
    \centering
    \begin{subfigure}[b]{0.245\textwidth}
        \centering
        \includegraphics[width=0.99\linewidth]{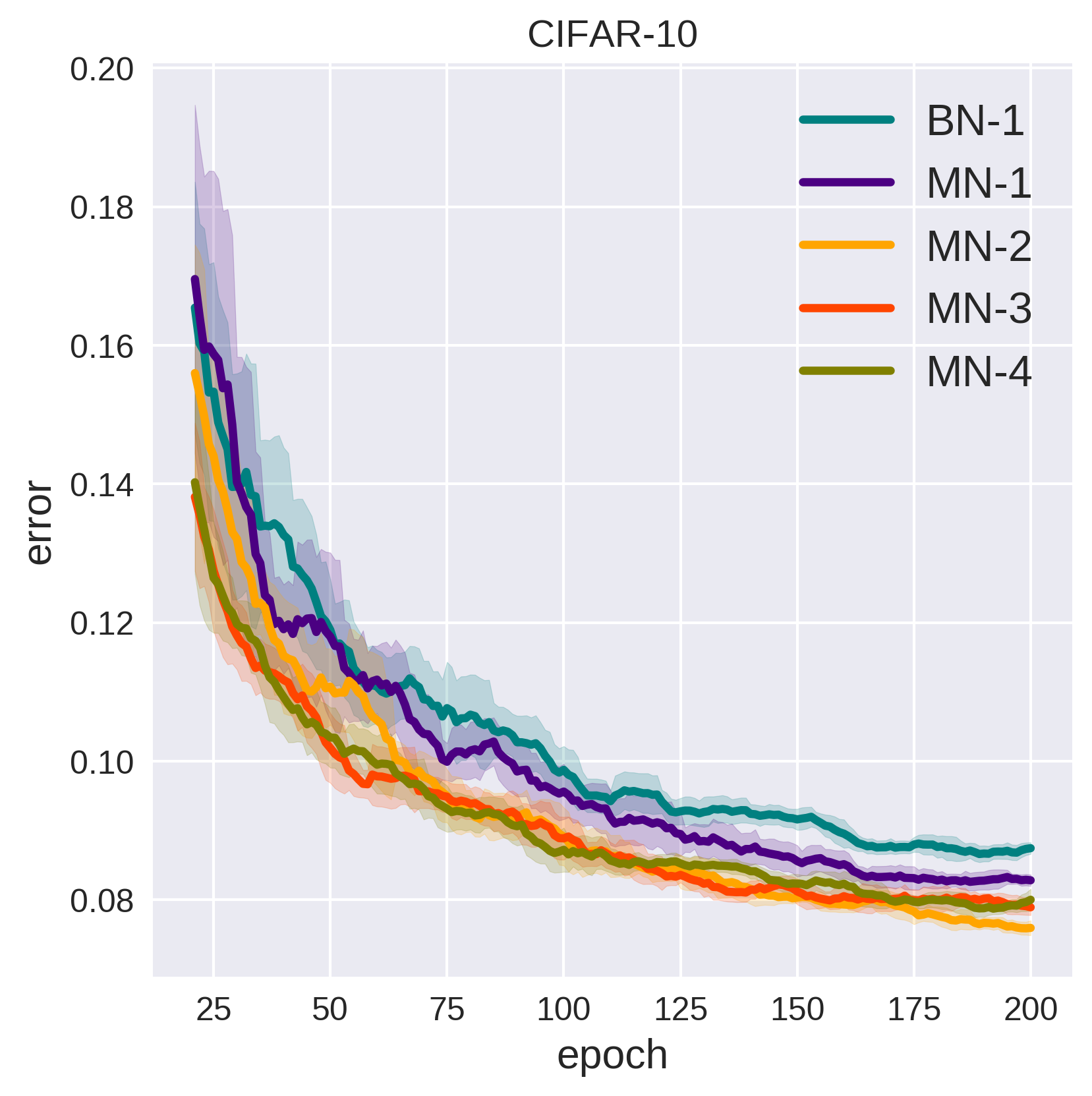}%
        \caption{}
        \label{fig:inceptionv3_cifar_10}
    \end{subfigure}
    \begin{subfigure}[b]{0.245\textwidth}
        \centering
        \includegraphics[width=0.99\linewidth]{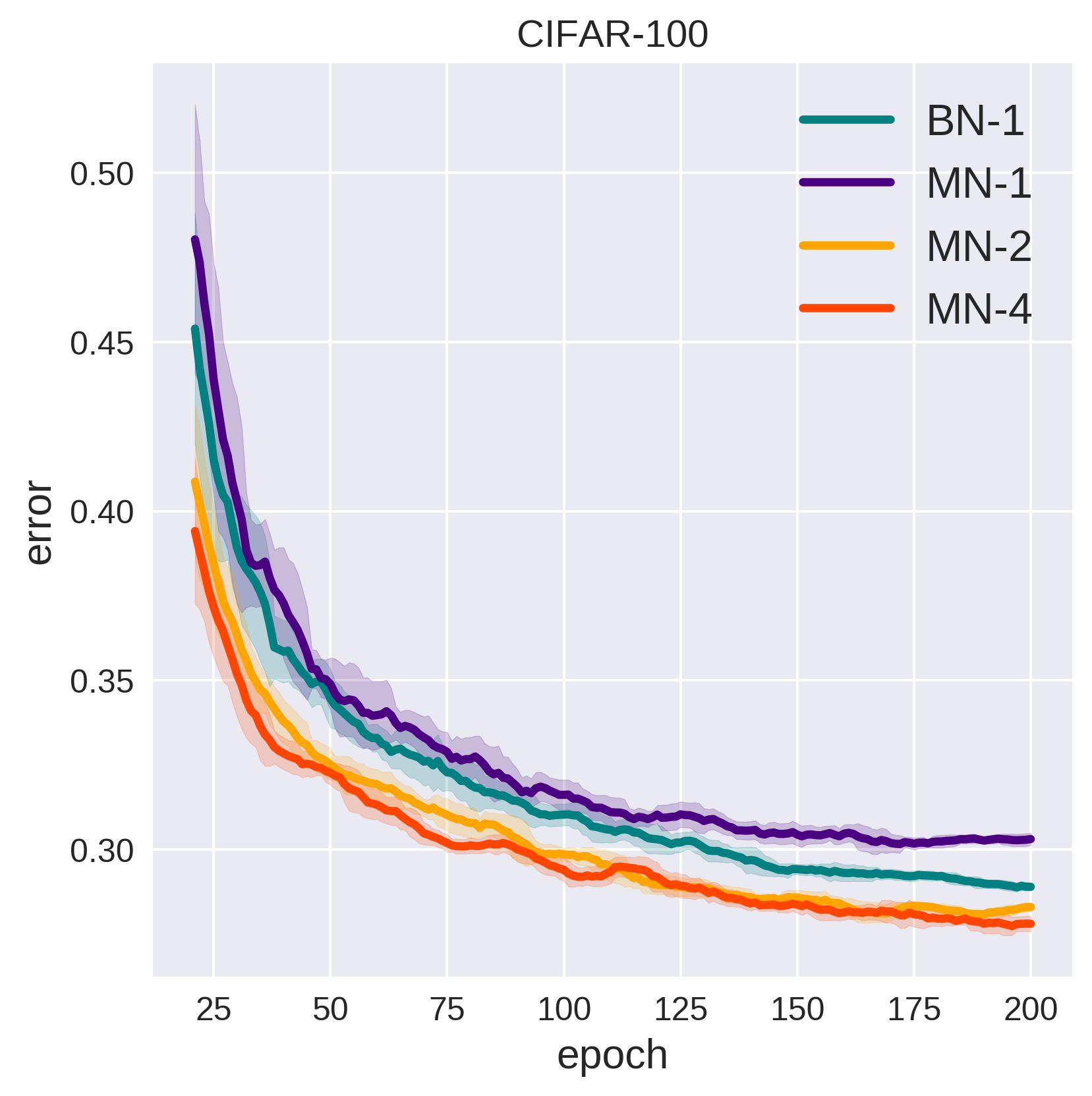}%
        \caption{}
        \label{fig:inceptionv3_cifar_100}
    \end{subfigure}
    \begin{subfigure}[b]{0.245\textwidth}
        \centering
        \includegraphics[width=0.99\linewidth]{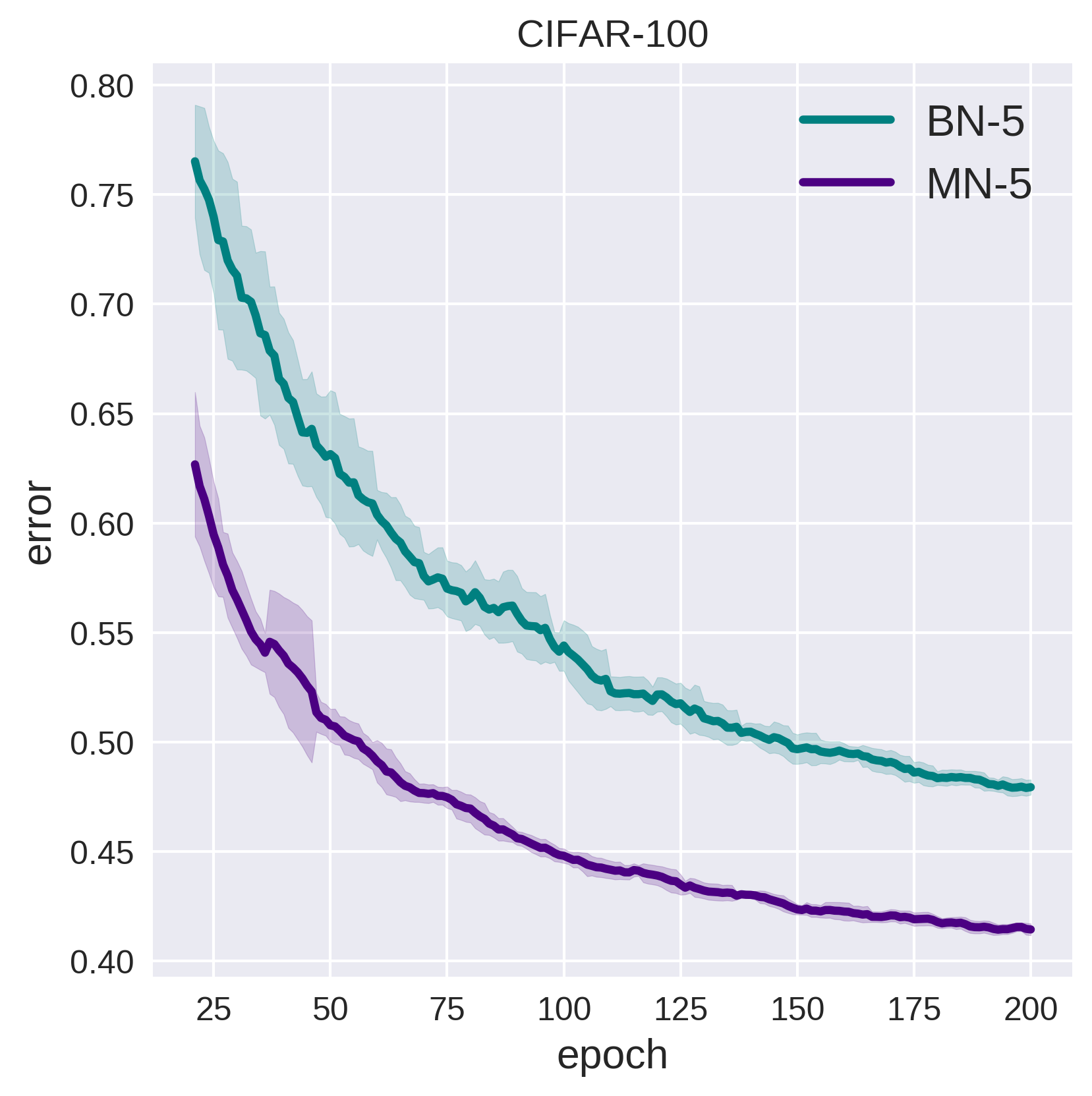}%
        \caption{}
        \label{fig:inceptionv3_cifar_100_large_lr}
    \end{subfigure}
    \begin{subfigure}[b]{0.245\textwidth}
        \includegraphics[width=0.99\textwidth]{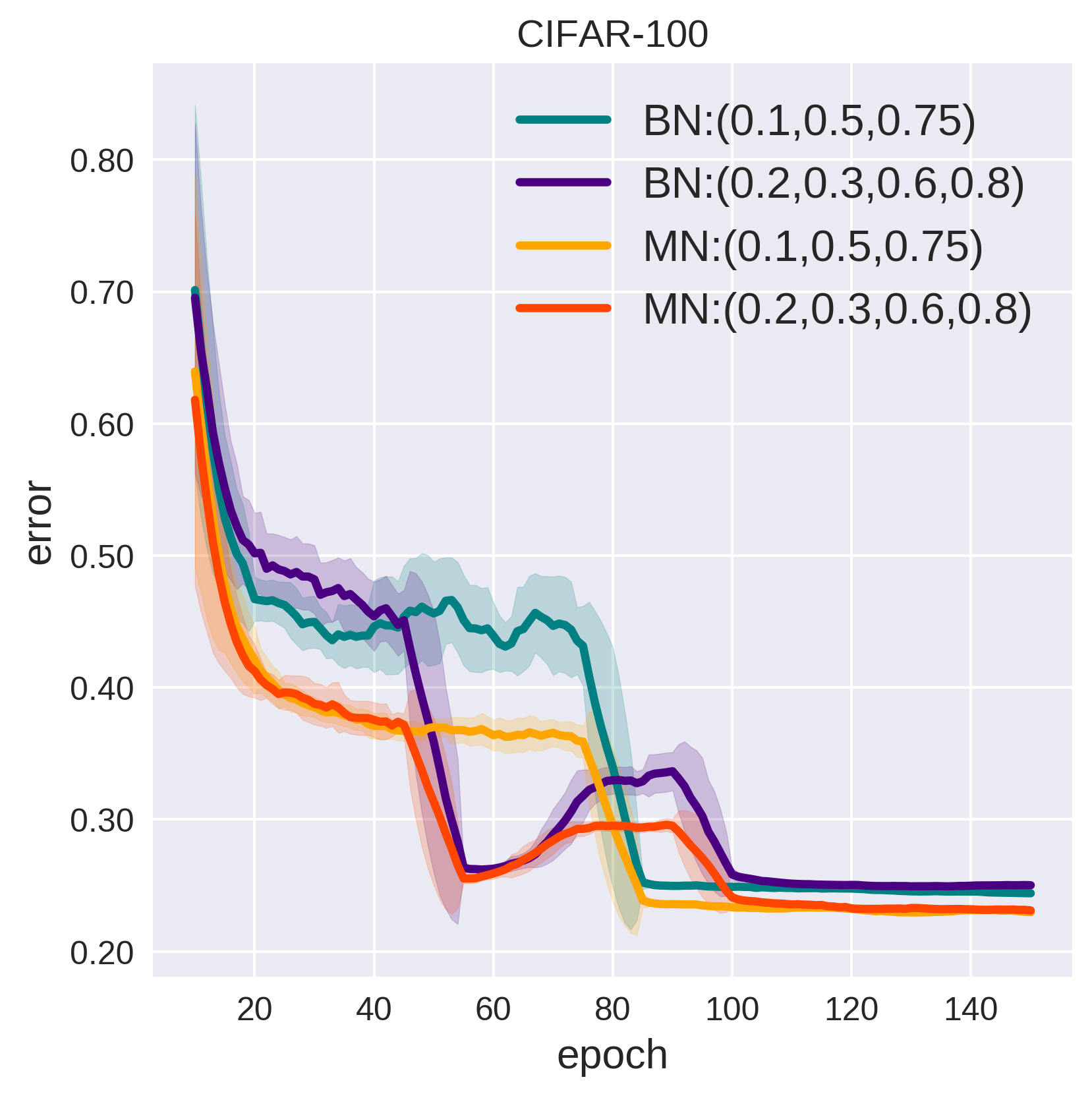}
        \caption{}
        \label{fig:nesterov_cifar100}    
    \end{subfigure}
    \caption{Test error curves when Inception-V3 architecture is trained under different settings. Figures \ref{fig:inceptionv3_cifar_10} and \ref{fig:inceptionv3_cifar_100} show the small learning rate regime, respectively, on CIFAR-10 and CIFAR-100. Figure \ref{fig:inceptionv3_cifar_100_large_lr} shows the large learning rate regime on CIFAR-100. Figure \ref{fig:nesterov_cifar100} illustrates test error curves of CIFAR-100 when Inception-V3 architecture is trained using Nesterov's accelerated gradient \cite{bengio2013advances} (all other experiments use RMSprop \cite{tieleman2012lecture}), with two different learning rate drop policies. Mixture normalization modules have been employed in ``inc2/1'' and ``inc3/0'' layers. We observe that across a variety of choices such as the number of mixture components, number of EM iterations, learning rate regime and drop policy, optimization technique, and the layer where MN is applied, mixture normalized models consistently accelerate their batch normalized counterparts and achieve better final test accuracy.}\label{fig:cifar_inceptionv3_results}
\end{figure*}

\begin{table}
  \centering
  \begin{tabular}{l*{2}{c}}
	\toprule
	\multicolumn{3}{c}{\textbf{CIFAR-10}}\\
	\midrule  
    model & steps to 91.50\% & max. acc.(\%)\\
    \midrule
    BN-1 & 7.34 $\times$ $10^4$ & 91.50\\
    MN-1 & 4.29 $\times$ $10^4$ & 91.91\\
    MN-2 & 3.86 $\times$ $10^4$ & 92.55\\
    MN-3 & 3.82 $\times$ $10^4$ & 92.30\\
    MN-4 & 3.58 $\times$ $10^4$ & 92.25\\
	\toprule
	\multicolumn{3}{c}{\textbf{CIFAR-100}}\\
	\midrule  
    model & steps to 71.30\% & max. acc.(\%)\\
    \midrule
    BN-1 & 7.34 $\times$ $10^4$ & 71.30\\
    MN-1 & -- & 70.43\\
    MN-2 & 4.57 $\times$ $10^4$ & 72.23\\
    MN-4 & 3.93 $\times$ $10^4$ & 72.74\\
    \bottomrule
  \end{tabular}
  \caption{For batch normalization and the mixture-normalized
variants using Inception-V3, the number of training steps required to
reach the maximum accuracy of batch-normalized model, and the maximum accuracy achieved by each variant.}
  \label{tab:cifar_inceptionv3_stepsto}
\end{table}

\begin{table}
  \centering
  \begin{tabular}{l*{2}{c}}
	\toprule
	\multicolumn{3}{c}{\textbf{lr decay policy: (0.1,0.5,0.75)}}\\
	\midrule  
    model & steps to 75.75\% & max. acc.(\%)\\
    \midrule
    BN & 2.78 $\times$ $10^4$ & 75.75\\
    MN & 1.52 $\times$ $10^4$ & 77.30\\
	\toprule
	\multicolumn{3}{c}{\textbf{lr decay policy: (0.2,0.3,0.6,0.8)}}\\
	\midrule  
    model & steps to 75.20\% & max. acc.(\%)\\
    \midrule
    BN & 2.41 $\times$ $10^4$ & 75.20\\
    MN & 1.79 $\times$ $10^4$ & 77.32\\
    \bottomrule
  \end{tabular}
  \caption{Training Inception-V3 using Nesterov's accelerated gradient \cite{bengio2013advances} on CIFAR-100, the number of training steps required to
reach the maximum accuracy of batch-normalized model along with the maximum accuracy achieved by each model.}
  \label{tab:nesterov_cifar100}
\end{table}

Table \ref{tab:cifar_inceptionv3_results} and Figure \ref{fig:cifar_inceptionv3_results} compare our proposed mixture normalization with its standard batch normalization counterparts. We observe that, with the exception of MN-1 on CIFAR-100, all the MN variants are not only successfully accelerating BN variants, but also achieve superior final test accuracy. As expected, and similar to the shallow network case, there is a benefit in using mixture normalization at multiple layers in different depth. Table \ref{tab:cifar_inceptionv3_results} also recommends using mixture normalization in later layers, closer to the classifier, in order to reach a better convergence rate and final test accuracy. 

\textbf{Effect of parameters:} We can observe from MN-3 versus MN-4 on CIFAR-10, that mixture normalization's performance is not sensitive to the number of mixture components assuming K is large enough. We have discussed above that the number of EM iterations directly affect the computation time. However, Table \ref{tab:cifar_inceptionv3_results} and Figure \ref{fig:cifar_inceptionv3_results}, demonstrate that by employing mixture normalization even with 2 EM updates, training a modern 48-layers deep architecture such as Inception-V3 will enjoy decent acceleration.

\textbf{Large learning rate:} To analyze the behavior of mixture normalization in large learning rate regime, we experiment training our models with the learning rate of 0.005 (5$\times$0.001). Figure \ref{fig:inceptionv3_cifar_100_large_lr} illustrates that MN-5 handles large learning rate better than BN-5, its batch normalized counterpart. Note that, to probe the sole effect of increasing the learning rate, here we keep the same learning rate decay policy (decay rate of 0.93 every four epochs) as the small learning rate regime. 

\textbf{Quantifying Acceleration:} Similar to the case of CIFAR CNN architecture, we report the number of steps (gradient descent updates) mixture normalization variants require in order to reach the maximum test accuracy obtained by their batch-normalized counterparts. Table \ref{tab:cifar_inceptionv3_stepsto} shows that the best performing variants of mixture normalization, MN-2 on CIFAR-10 and MN-4 on CIFAR-100, reduce the number of training steps towards 91.50\% and 71.30\% test accuracy, respectively on CIFAR-10 and CIFAR-100, by $\sim$\textbf{47}\%. Similar to the case of shallow CIFAR CNN architecture, here, mixture normalization, also improves the final test accuracy by $\sim$1-1.5\%.

\textbf{Choice of optimization technique:} We evaluate whether the performance of mixture normalization is consistent using different choices of optimization technique. That is, instead of RMSprop \cite{tieleman2012lecture}, we use Nesterov's accelerated gradient \cite{bengio2013advances}, with 0.9 as momentum and decay the learning rate in two different fashions. First, following the policy adopted in ResNet \cite{he2016deep} and DenseNet \cite{huang2017densely}, we reduce the learning rate twice by a factor of 10 at 50\% and 75\% of the total number of epochs. Second, following Wide Residual networks \cite{zagoruyko2016wide}, we reduce the learning rate three times with a factor of 5 at 30\%, 60\% and 80\% of the total number of epochs. For MN models, mixture normalization is employed at ``inc2/1'' and ``inc3/0'' layers. Mini-batch size and the initial learning rate are respectively set to 256 and 0.14 and we train all the models for 150 epochs. 

Figure \ref{fig:nesterov_cifar100} shows the test error curves on CIFAR-100. Similar to the previous experiments, in Table \ref{tab:nesterov_cifar100}, we compare the required number of gradient updates in order to reach the maximum test accuracy of batch normalized model. In both learning rate decay scenarios, mixture normalization is not only able to considerably accelerate training procedure, but also achieves $\sim$2\% better final test accuracy. Finally, in comparison with the state-of-the-art architectures of comparable depth, our MN model, with a test error of \textbf{22.68}\% on CIFAR-100, performs on par with Wide Residual networks \cite{zagoruyko2016wide} while outperforming DenseNet \cite{huang2017densely} (ref. Table 2 in \cite{huang2017densely}).

\begin{figure*}[h]
    \centering
    \begin{subfigure}[b]{0.245\textwidth}
        \centering
        \includegraphics[width=0.99\linewidth]{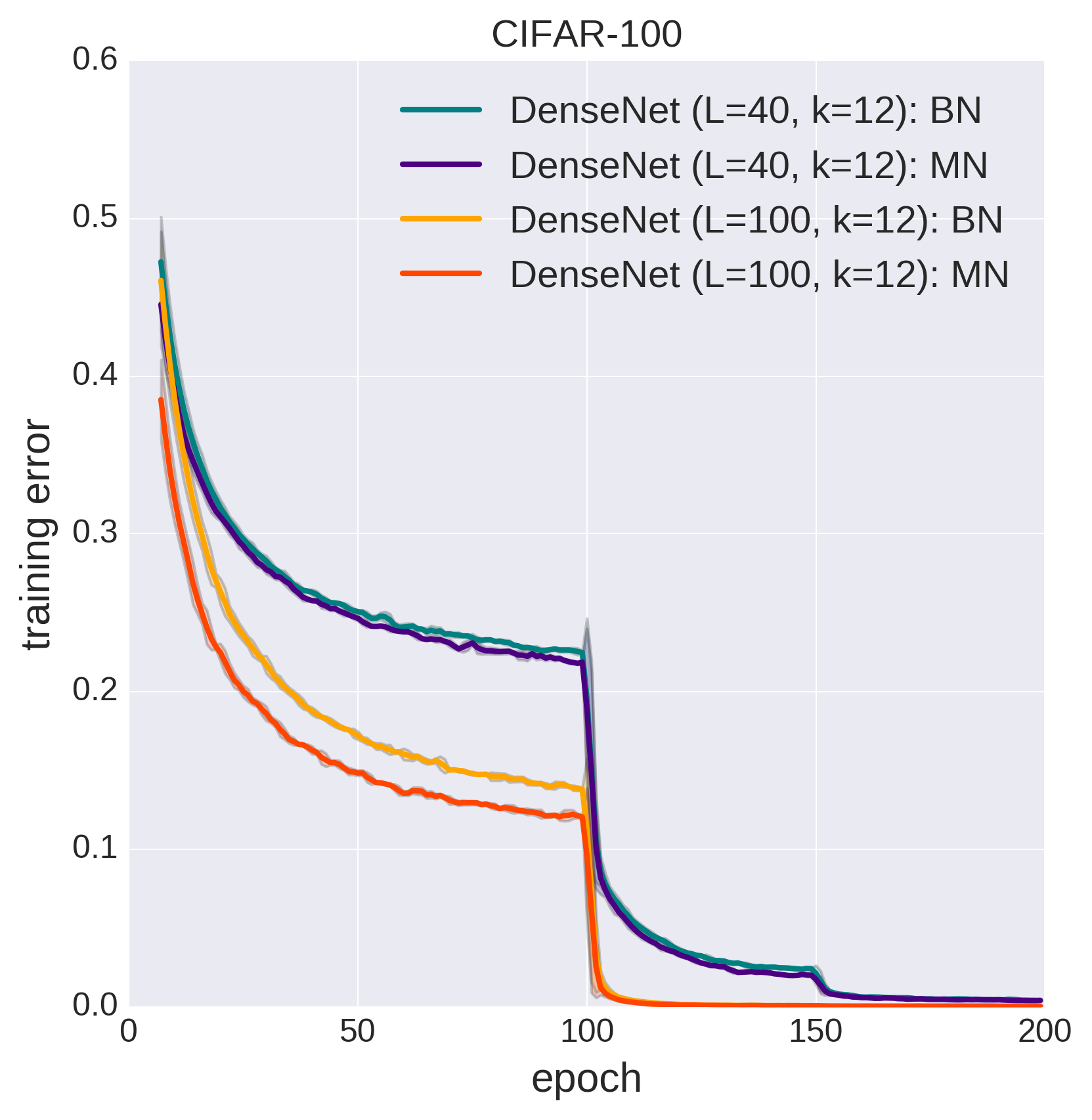}%
        \caption{}
        \label{fig:DenseNet_train_error}
    \end{subfigure}
    \begin{subfigure}[b]{0.245\textwidth}
        \centering
        \includegraphics[width=0.99\linewidth]{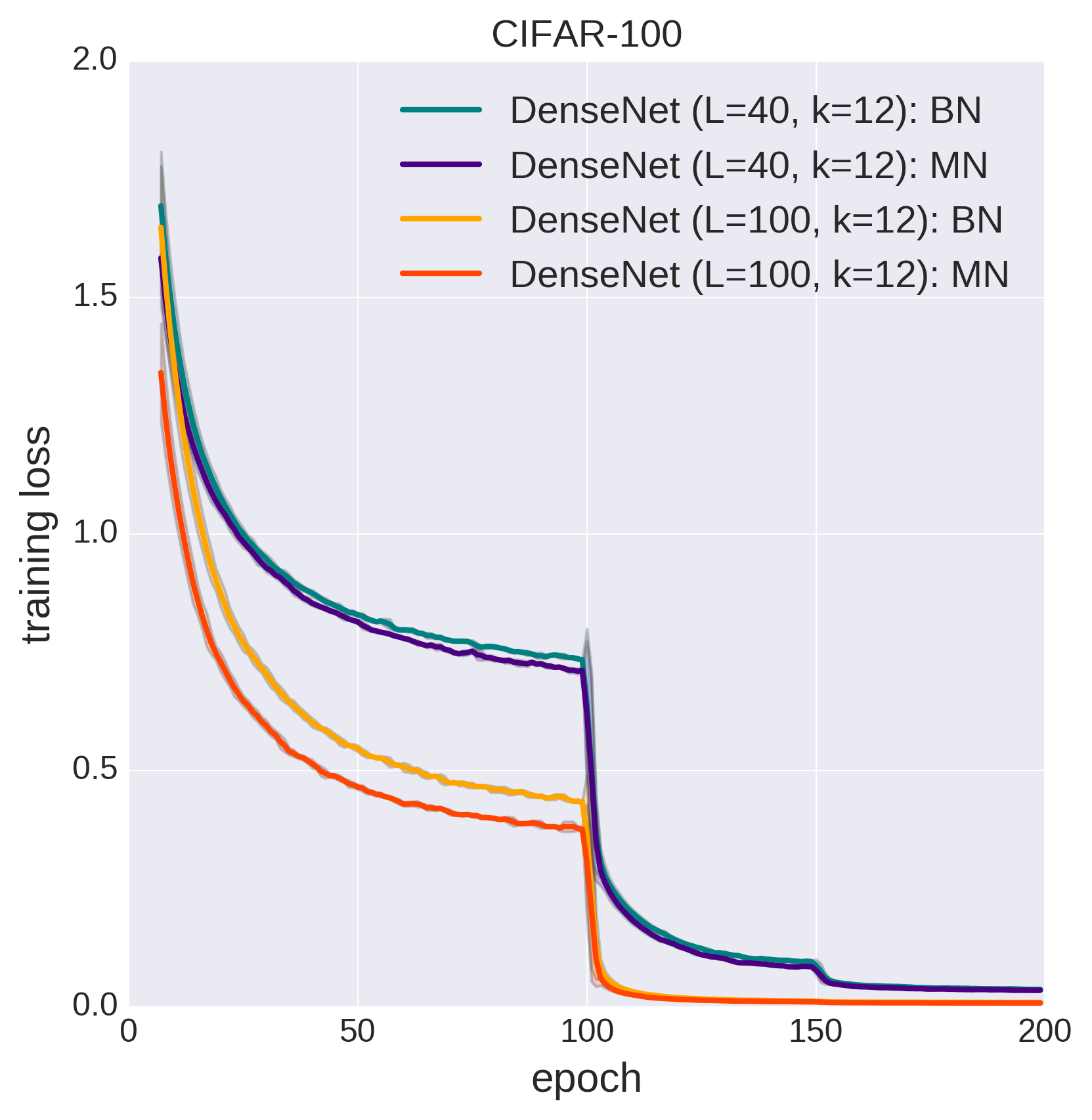}%
        \caption{}
        \label{fig:DenseNet_train_loss}
    \end{subfigure}
    \begin{subfigure}[b]{0.245\textwidth}
        \centering
        \includegraphics[width=0.99\linewidth]{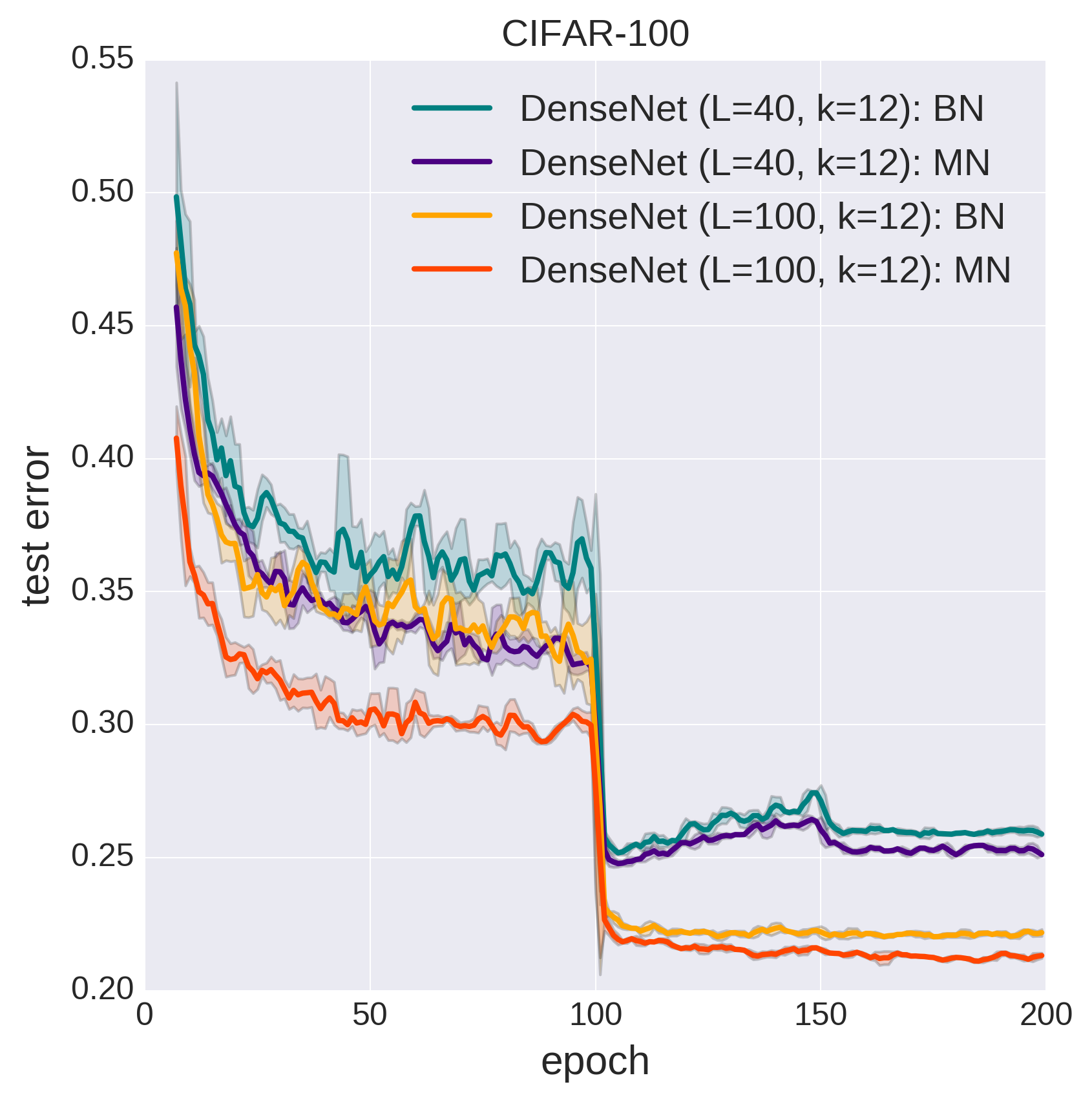}%
        \caption{}
        \label{fig:DenseNet_test_error}
    \end{subfigure}
    \begin{subfigure}[b]{0.245\textwidth}
        \includegraphics[width=0.99\textwidth]{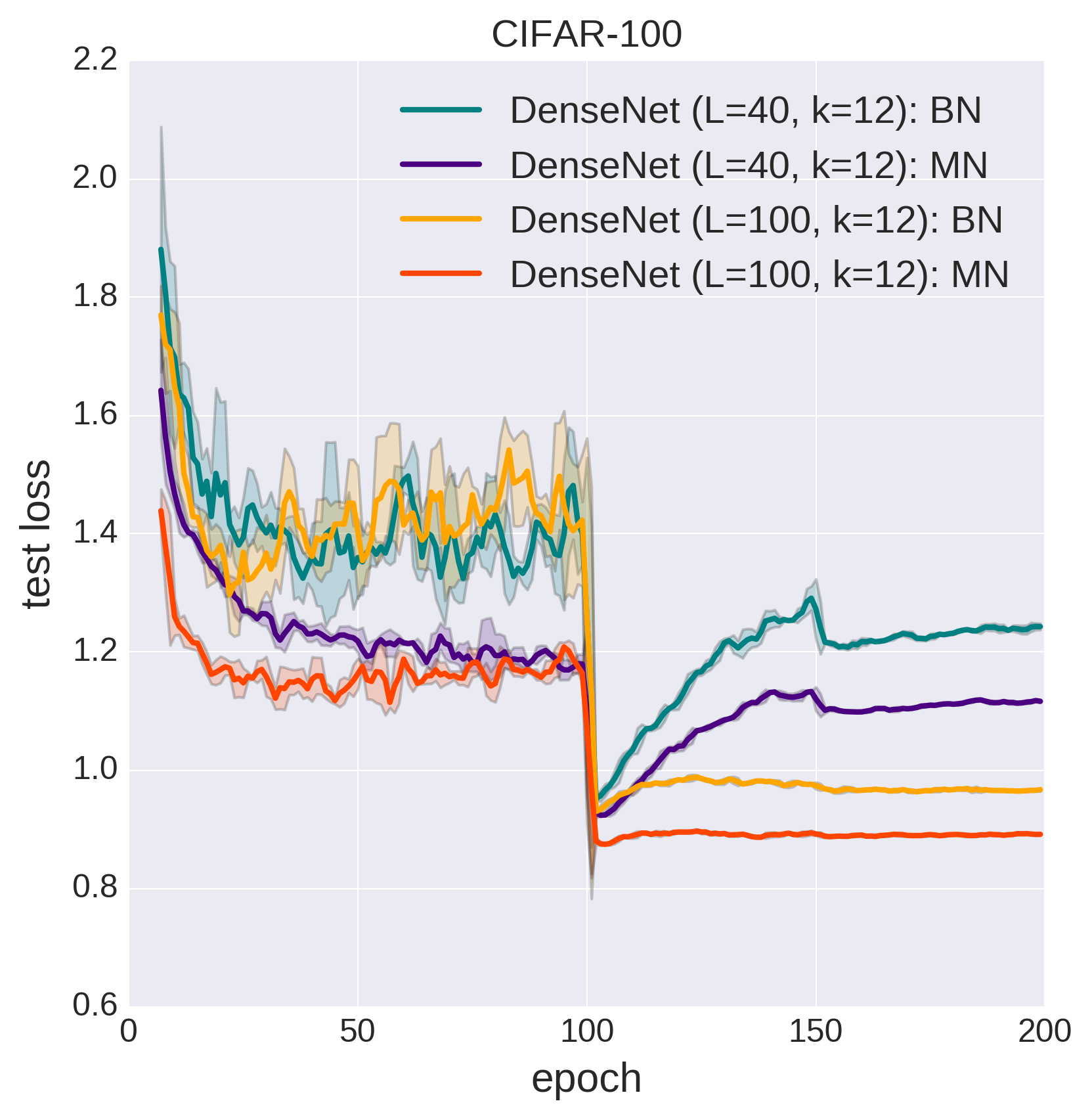}
        \caption{}
        \label{fig:DenseNet_test_loss}    
    \end{subfigure}
    \caption{\hlcyan{DenseNet \cite{huang2017densely} experiments on CIFAR-100. Figures \ref{fig:DenseNet_train_error} and \ref{fig:DenseNet_train_loss} respectively illustrate the training error and cross entropy loss. Figures \ref{fig:DenseNet_test_error} and \ref{fig:DenseNet_test_loss} respectively illustrate the test error and cross entropy loss. We observe from \ref{fig:DenseNet_train_error} and \ref{fig:DenseNet_train_loss} that mixture normalization facilitates the training process by accelerating the optimization. Meanwhile it provides better generalization (ref. \ref{fig:DenseNet_test_error}, and \ref{fig:DenseNet_test_loss}) by continuously maintaining a large gap with respect to its batch normalized counterpart. We show
    1 standard deviation (shaded area) computed within a window of 3 epochs for all the curves.}}\label{fig:cifar_densenet}
\end{figure*}

\hlcyan{
\subsection{DenseNet}\label{sec:cifar_densenet}
We conclude our experimental results by evaluating the effectiveness of mixture normalization in DenseNet \mbox{\cite{huang2017densely}} architectures. We use two basic architectures with 40 and 100 layers. There are three dense blocks and two transition layers. Both architectures are using a growth rate of 12. All models are trained for 200 epochs on CIFAR-100, with the batch size of 64 using Nesterov's accelerated gradient \mbox{\cite{bengio2013advances}}. Learning rate is initialized at 0.1 and is divided by 10 at 50\% and 75\% of the total number of training epochs. We set the weight decay and momentum values to $10^{-4}$ and 0.9, respectively. Figure \mbox{\ref{fig:cifar_densenet}} illustrates the classification error and cross entropy loss of training and test. For mixture normalized models, denoted by MN, we solely replace the batch normalization layers of two transition layers and the last (after the third dense block), with mixture normalization. The number of components and EM iterations for MN variants are set to 5 and 2, respectively. From Figure \mbox{\ref{fig:cifar_densenet}}, we observe that mixture normalization not only facilitates training by accelerating the optimization, but also consistently provides better generalization on both architecture settings. In fact, the benefit of mixture normalization is more clear on the deeper setting of the DenseNet architecture. Using DenseNet \{L=40, k=12\} on CIFAR-100, the best test error achieved by the BN variant is \textbf{25.10}\% versus \textbf{24.63}\% of its MN counterpart. When we switch to deeper DenseNet \{L=100, k=12\}, architecture, this number reduces to \textbf{21.93}\% for BN-based model while MN variant reaches the best test error of \textbf{20.97}\%.}

\subsection{Mixture Normalization in GANs}\label{sec:mixturenorm_in_gan}
Generative adversarial networks (GANs) \cite{goodfellow2014generative} have recently shown amazing progress in generating new realistic images \cite{radford2015unsupervised}\cite{ledig2016photo}\cite{isola2017image}\cite{arjovsky2017wasserstein}. The training process is a minimax game between generator and discriminator. Discriminator learns to separate real images from the fake ones, created by the generator. Meanwhile, generator attempts to impede discriminator's job by progressively generating more realistic images.
Hence, at convergence, generator, theoretically, must be able to generate images, whose distribution matches the distribution of the real images. However, in practice, ``mode collapse''\cite{arjovsky2017towards}\cite{che2016mode}\cite{chen2016infogan}\cite{metz2016unrolled}\cite{salimans2016improved}\cite{ghosh2017multi} problem prevents generator from learning a diverse set of modes with high probability, as is in the distribution of real images. Therefore, since our proposed mixture normalization, normalizes internal activations, independently over multiple disentangled modes of variation, we hypothesize that employing it in generator, should improve the training procedure of GANs.

To evaluate our hypothesis, we consider popular Deep Convolutional GAN (DCGAN) \cite{radford2015unsupervised} architecture. Its generator consist of one linear and four deconvolution layers. The first three deconvolution layers are separated from each other by batch normalization \cite{ioffe2015batch} followed by ReLU\cite{nair2010rectified} activation function. We replace the batch normalization layers associated with the first two deconvolution layers with mixture normalization (K=3, EM iter=2). We train all the models on CIFAR-10 for 100K updates (iterations) using Adam \cite{kingma2014adam} with $\alpha=0.0002$, $\beta_1=0$ and $\beta_2=0.9$ for both generator and discriminator. The quality of GANs are measured using ``Fr\'echet Inception Distance''(FID) \cite{heusel2017gans}, evaluated every 10K updates for computational efficiency.

\begin{figure}[htbp]
    \centering
    \begin{subfigure}[b]{0.5\textwidth}
        \includegraphics[width=0.8\textwidth]{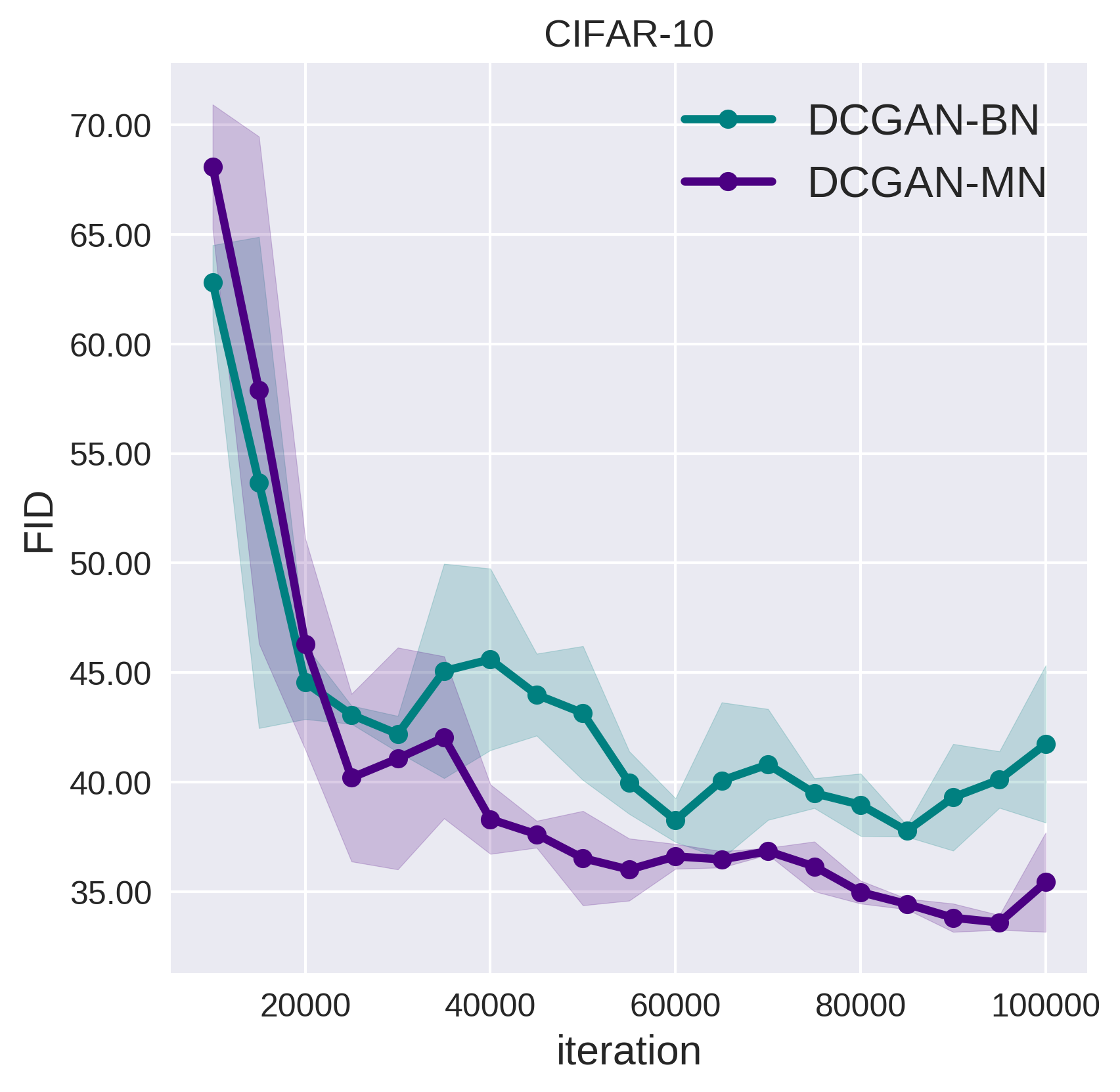}
    \end{subfigure}
    \caption{Mixture normalization in deep convolutional GAN (DCGAN)\cite{radford2015unsupervised}. We observe that employing our proposed mixture normalization in the generator of DCGAN (DCGAN-MN) facilitates the training process. In comparison with the standard DCGAN which uses batch normalization (DCGAN-BN), DCGAN-MN not only converges faster (a reduction of $\sim$58\%) but also achieves better (lower) FID (33.35 versus 37.56). For better visualization, we show one standard deviation (shaded area) computed within a window of 30K iterations (3 adjacent FID evaluation points).}\label{fig:dcgan_curve}
\end{figure}

\begin{figure}[h]
    \centering
    \begin{subfigure}[b]{0.5\textwidth}
        \centering
        \includegraphics[width=0.8\linewidth]{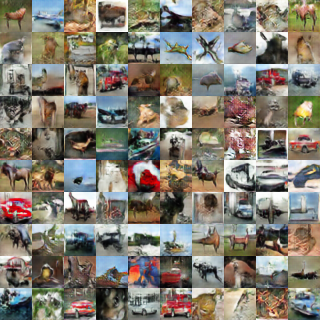}%
        \caption{DCGAN-BN}
        \label{fig:bngan_image}
    \end{subfigure}
    \begin{subfigure}[b]{0.5\textwidth}
        \centering
        \includegraphics[width=0.8\linewidth]{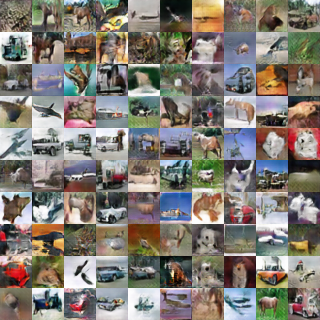}%
        \caption{DCGAN-MN}
        \label{fig:mngan_image}
    \end{subfigure}
    \caption{Samples of generated images by batch and mixture normalized DCGAN models, at their best (lowest) evaluated FID, are respectively illustrated in Figure \ref{fig:bngan_image} and \ref{fig:mngan_image}. DCGAN-BN and DCGAN-MN, respectively achieve FID of 37.56 and 33.35.}\label{fig:gan_image}
\end{figure}

Figure \ref{fig:dcgan_curve} demonstrates that mixture normalized DCGAN (DCGAN-MN) not only converges faster than its batch normalized counterpart but also achieves better (lower) FID. While DCGAN-BN reaches the lowest FID of 37.56 (very close to 37.7 reported in \cite{heusel2017gans}) after 60K steps (gradient updates). It only takes 25K steps, a reduction of $\sim$58\%, for DCGAN-MN to reach 37.56. Furthermore, the lowest FID obtained by DCGAN-MN is 33.35, a significant improvement over the batch normalized model. Figure \ref{fig:gan_image} illustrates samples of generated images by batch and mixture normalized DCGANs at their lowest evaluated FID.

\section{Computational Complexity and Detailed Analysis}\label{sec:discussion}
In this section, we give an in-depth analysis on the behavior of the mixture normalization. Specifically, we provide computational complexity analysis, both in theory and practice. We discuss how representation of mixture components evolves and whether the rationale to utilize mixture model remains valid as we pass early stages of the training process. 

\hlcyan{
\subsection{Computational Complexity Analysis}\label{sec:complexity}
The computational overhead of mixture normalization, in comparison to the batch normalization \mbox{\cite{ioffe2015batch}}, is in estimating the parameters of the Gaussian mixture model. This is a two-stage process where, we use a seeding procedure to initialize the centers of the mixture components, followed by estimating the parameters of the mixture model through iterations of Expectation-Maximization (EM).

In our implementation\footnote{\hlcyan{Mixture Normalization is implemented in Chainer \cite{chainer_learningsys2015}, and thanks to CUPY \cite{cupy_learningsys2017}, the entire computation including K-means++ and EM iterations are performed in GPU.}}, we use K-means++ \mbox{\cite{arthur2007k}} as the seeding procedure. Without any assumption on the data, it is in expectation $O(\log k)$-competitive with the optimal K-means clustering solution \mbox{\cite{arthur2007k}}, and has a complexity of $\Theta(nkd)$ where, $n$ is the number of data points, $k$ is the number of cluster centers and $d$ is the dimensionality of the data \mbox{\cite{bachem2016approximate}}. However, there is a rich literature on algorithms that speed up K-means++ \mbox{\cite{arthur2007k}} seeding procedure. For example, Bachem \textit{et. al} \mbox{\cite{bachem2016approximate}} have proposed a MCMC-based sampler to approximate the full seeding step of K-means++ with complexity of $\Theta(mk^{2}d)$. Under
some light assumptions, the authors prove that their proposed solution is in expectation $O(\log k)$-competitive with the optimal solution, if we have the chain length $m\in \Theta(k\log^{2}n \log k)$. In this case, the total computational
complexity will be $O(k^{3}dn\log k)$, which is sublinear in the number of data points. Authors in \mbox{\cite{bachem2016fast}} show that an assumption-free version of \mbox{\cite{bachem2016approximate}} with lower complexity is also achievable. Finally, to ameliorate the inherent sequential passes over data, Scalable K-means++ \mbox{\cite{bahmani2012scalable}} with nice theoretical support \mbox{\cite{bachem2017distributed}} can be used. These all indicate that the initial seeding procedure used in mixture normalization can be implemented with low computational complexity while maintaining provably good seeding. It is easy to show that assuming diagonal covariance, each EM iteration to fit the GMM parameters is of complexity $O(nkd)$ \mbox{\cite{verbeek2003efficient}}, where $n$ can significantly be reduced using coresets \mbox{\cite{lucic2017training}}. In practice, we found that one does not need to use all the data points to estimate the parameters of the mixture model. Instead a simple random sampling that maintains at least 25\% of the data points is sufficiently accurate. Hence, we confirm that estimating the parameters of the Gaussian mixture model can be done efficiently in scale.

In order to see how the above complexity analysis translates in practice, we compared the computation cost of mixture normalization against native implementation of batch normalization \cite{ioffe2015batch}. Experiments are conducted on CIFAR-100 using Densenet \mbox{\cite{huang2017densely}} architecture where, MN replaces the last (after the third dense block) BN layer. We tried Densenet (3 blocks and growth rate of 12) with 20 and 40 layers which result in the number of input channels to batch/mixture normalization to be respectively 196 and 448. This allows us to evaluate the effect of dimensionality of data points. We also used batch sizes of 64 and 128 in order to study the effect of the number of data points. Finally, we varied the number of mixture components ($K$) from 2 to 5 to analyze its effect on the computation cost. Note that the iteration/sec. counts for the computation of the entire network not just the batch/mixture normalization layer. Table \mbox{\ref{tab:computation_cost}} indicates that in practice, the computation cost of MN scales very well with respect to the number of datapoints, dimensionality of data and the number of mixture components.}

\begin{table}
  \centering
  \begin{tabular}{l*{4}{c}}
	\toprule
    model & K & batch size & dim. & iteration/sec.\\
    \midrule
    BN & 1 & 64 & 196 & 16.40\\
    MN & 2 & 64 & 196 & 14.46\\
    MN & 3 & 64 & 196 & 14.42\\
    MN & 4 & 64 & 196 & 13.91\\
    MN & 5 & 64 & 196 & 13.69\\
    \midrule
    BN & 1 & 64 & 448 & 9.26\\
    MN & 2 & 64 & 448 & 8.83\\
    MN & 3 & 64 & 448 & 8.51\\
    MN & 4 & 64 & 448 & 8.52\\
    MN & 5 & 64 & 448 & 8.41\\
    \midrule
    BN & 1 & 128 & 448 & 7.35\\
    MN & 2 & 128 & 448 & 6.59\\
    MN & 3 & 128 & 448 & 6.51\\
    MN & 4 & 128 & 448 & 6.41\\
    MN & 5 & 128 & 448 & 6.38\\
    \bottomrule
  \end{tabular}
  \caption{\hlcyan{Computation cost comparison of mixture normalization against natively implemented (no CUDA-kernel) batch normalization \cite{ioffe2015batch}. Experiments are conducted on a Titan X (Pascal) GPU.}}
  \label{tab:computation_cost}
\end{table}

\subsection{Evolution of Mixture Components}\label{sec:evolution}
\begin{figure}[htbp]
    \centering
    \begin{subfigure}[b]{0.5\textwidth}
        \includegraphics[width=0.99\textwidth]{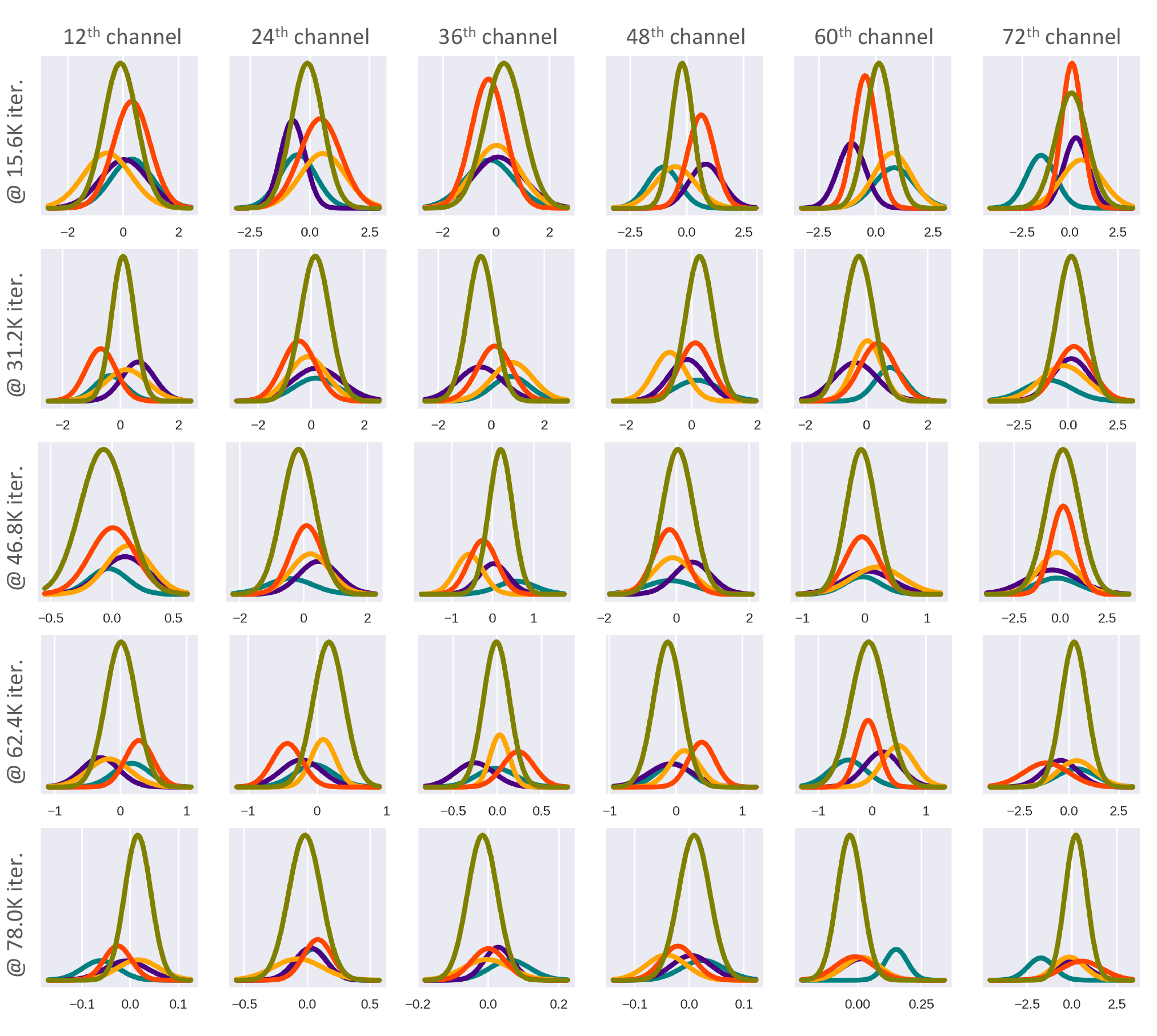}
    \end{subfigure}
    \caption{Evolution of mixture normalization associated to ``inc2/0/2/4'' layer as training MN-4 on CIFAR-100 progresses. As argued before, we observe that the underlying distribution is comprised of multiple modes of variation. While these sub-populations are of relatively uniform importance at the beginning, as training procedure goes on, mixture components evolve where some get closer, while others are pushed away from each other creating more distinct components. Here, different colors index mixture components, when sorted according to $\lambda_{k}$ values. }\label{fig:evolution}
\end{figure}

To better understand how mixture normalized models evolve, we visualize in Figure \ref{fig:evolution}, the mixture components associated to ``inc2/0/2/4''\footnote{It refers to when the fifth batch normalization in the third branch of the first inception layer in second inception block is replaced with mixture normalization.} layer as training MN-4 (ref. Table \ref{tab:cifar_inceptionv3_results}) on CIFAR-100 progresses. We show a random subset of 192 channels at 20\%, 40\%, 60\%, 80\% and 100\% of total training iterations. 

There are two main observations here. First, in early stages of training, multiples modes captured by different mixture components are of relatively similar weights ($\lambda_{k}$ in Equation \ref{eq:gmm0}). That is aligned with our argument that due to non-linearities, underlying distributions are comprised of multiple modes of variation. However, as training procedure goes on, mixture components evolve as some get closer, while others are pushed away from each other creating more distinct components. Second, notice how the horizontal axis, associated with the activation values, reduces in range. This alongside with distribution of $\lambda_{k}$s morphing from relatively uniform into one with a dominant bin (component in olive), demonstrates that mixture normalization, as intended, \textit{tries} to transform the underlying distribution of activations from a wide distribution comprised of multiple large components into a narrow one with a single dominant mode. Notice that despite a dominant component emerging, other components do not necessarily vanish rather their contribution diminishes. We will later show that such imbalance between $\lambda_{k}$ of various components is not large enough to trigger major mode collapse. These observations confirm that our hypothesis regarding the nature of the underlying distribution is valid and mixture normalization in practice, follows what its formulation is advocating.

\begin{figure}[htbp]
\begin{subfigure}{.5\textwidth}
  \centering
  \includegraphics[width=0.85\linewidth]{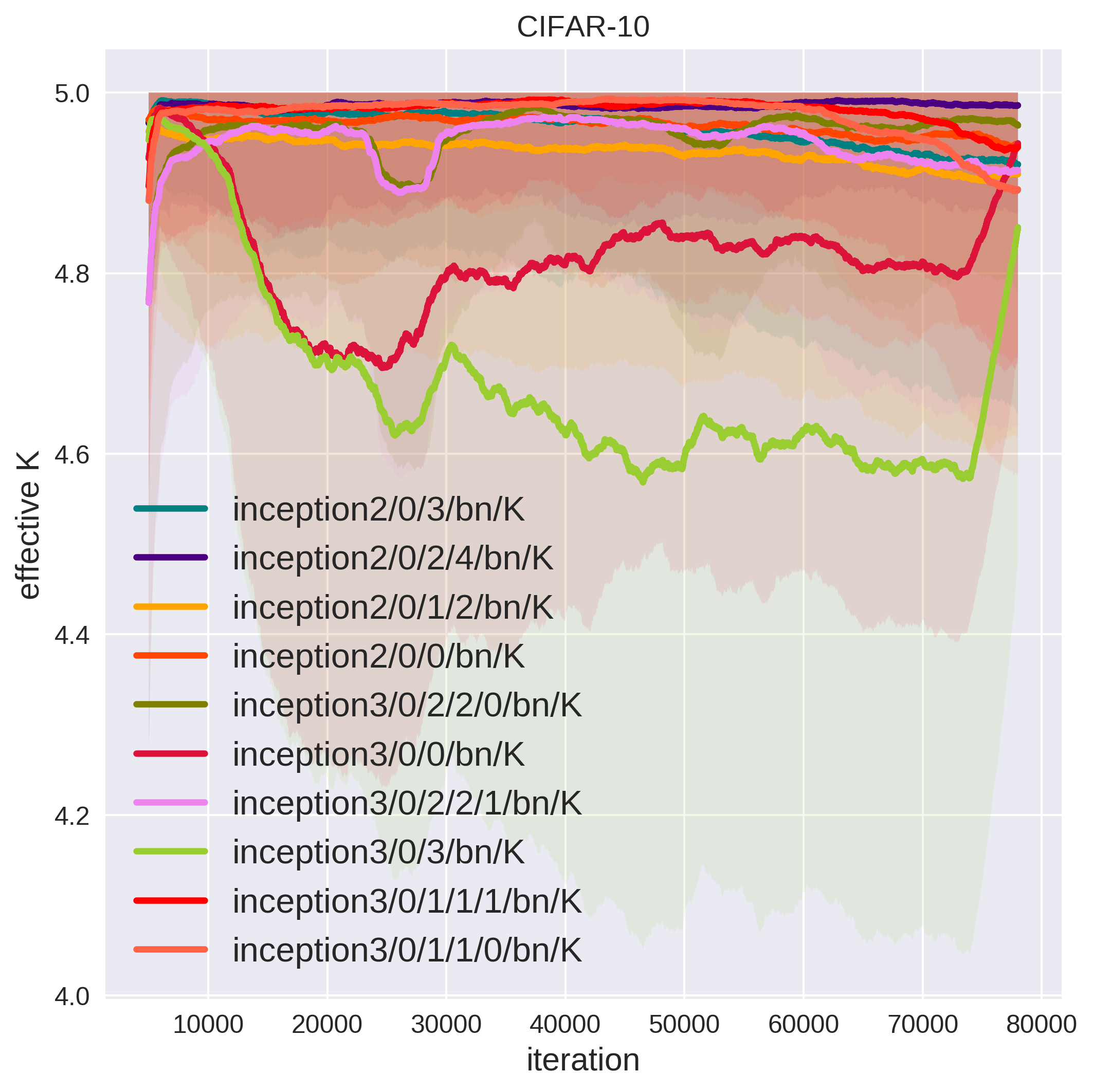}
  \vfill
  \includegraphics[width=0.85\linewidth]{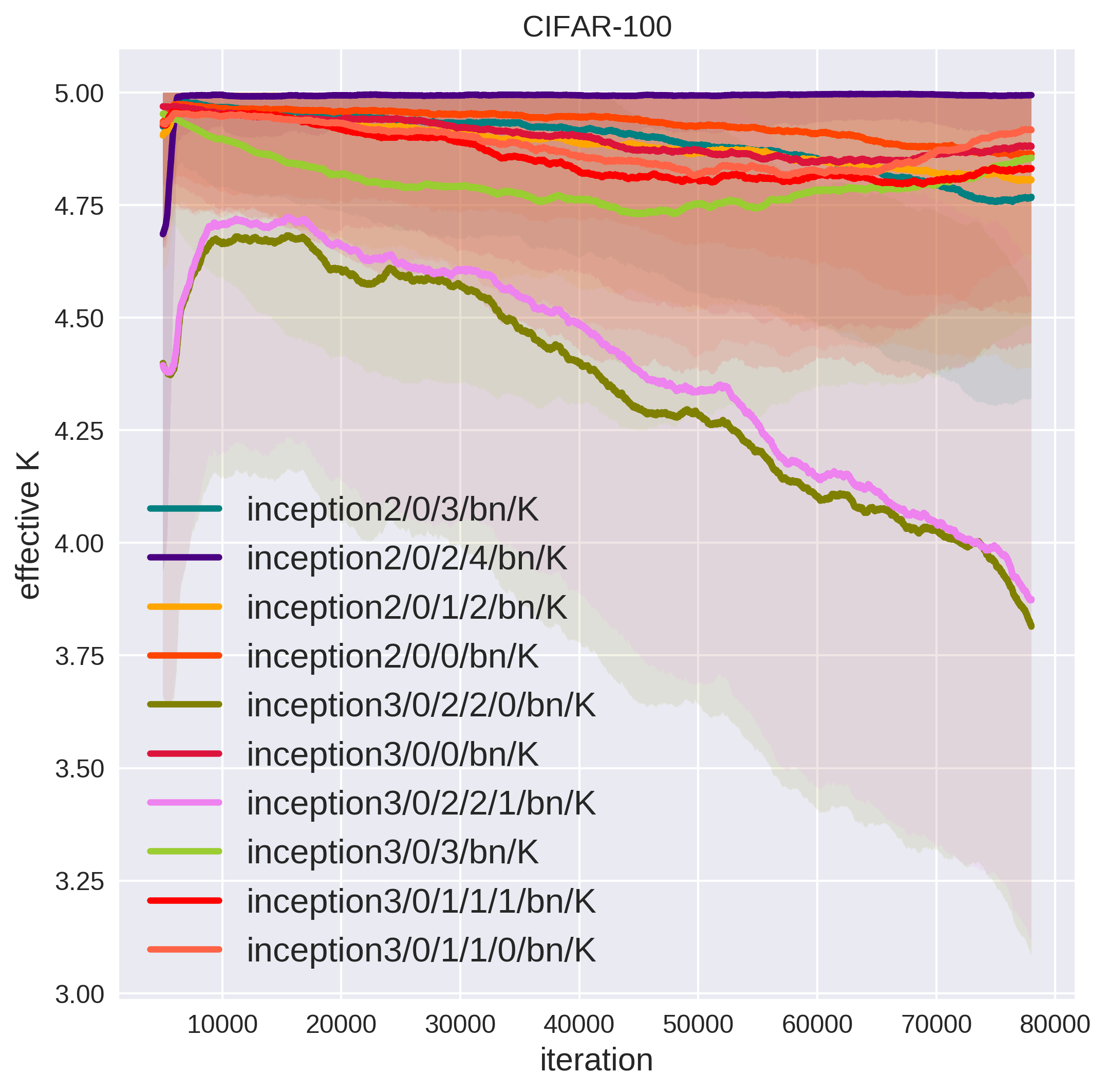}
\end{subfigure}
\caption{Effect of the number of mixture components in MN-4 using Inception-V3 when trained on CIFAR-10 and CIFAR-100. For the sake of better visualization, curves are smoothed using running average and one standard deviation is shown as the shaded area. We can see that the majority of MN modules fully utilize all ($K$=5) their mixture components, indicating that the need for better approximation using mixture model does not disappear, rather slightly diminishes, as the training procedure continues.}
\label{fig:effective_K}
\end{figure}

\subsection{Effective Number of Mixture Components}\label{sec:effective_K}
One of the hyperparameters of mixture normalization is the number of its components, K, which must be specified as a choice of design. In some cases, K-means clustering or GMM may generate components with very small $\lambda_{k}$, meaning that the corresponding component is not very representative. Therefore, we, in our implementation, have opted heuristics to discard components whose normalized $\lambda_{k}$ is less than 0.01, where samples associated to those are then merged with the remaining components. If the underlying data distribution is sufficiently well approximated using a single Gaussian distribution, which is against this paper's proposal, then we should expect mixture normalization to not utilize all the K components. Figure \ref{fig:effective_K} illustrates the actual number of components that mixture normalization has used through the entire training procedure of MN-4 (ref. Table \ref{tab:cifar_inceptionv3_results}) on CIFAR-10 and CIFAR-100. We show the curves associated to all the 10 mixture normalization modules in ``inc2/0'' and ``inc3/0''. We observe that except two cases, the rest of mixture normalization modules consistently utilize all the K=5 components. This suggests that despite the potential appearance of a dominant component (ref. Section \ref{sec:evolution}), properly estimating the underlying distribution of activations, still prefers a mixture model over a single Gaussian distribution.

\section{Conclusion}\label{sec:conclusion}
Batch normalization (BN) \cite{ioffe2015batch} uses mean and standard deviation, computed over entire instances within a mini-batch, in order to normalize input activations to the layers.
In this paper, we proposed improving BN by normalizing with respect to multiple mean and standard deviations associated with different modes of variation in the underlying data distribution. Specifically, each instance in the mini-batch, is softly assigned to a component of the Gaussian mixture model, where the GMM approximates the probability density function of input activations. Our proposed mixture normalization (MN) technique by piecewise re-structuring the probability density function of mini-batch distribution is able to drastically accelerate training deep CNN architectures. Experiments on CIFAR-10 and CIFAR-100 \cite{krizhevsky2009learning}, in a variety of training scenarios, using both shallow and very deep architectures, confirm the effectiveness of mixture normalization. We have provided detailed analysis on the evolution of mixture components as they go through training procedure. Beyond image classification, inspired by the ``mode collapse'' problem in training GANs \cite{goodfellow2014generative}, we have demonstrated that mixture normalization when integrated in the layers of generator, not only accelerates the training process of DCGAN \cite{radford2015unsupervised} but also results in higher quality models. Hence, we affirm that in the context of deep convolutional neural networks, underlying data distribution is indeed better approximated via GMM and training benefits from such mixture modeling that MN provides. In future, we are going to study the application of mixture normalization, to the recurrent neural networks. Furthermore, we will explore estimating Gaussian mixture model, via batch and stochastic Riemannian optimization \cite{hosseini2017alternative}. This would replace the EM optimization, that is currently performed outside the computational graph of the deep neural networks. It also allows us to utilize full covariance matrix to better approximate distribution of the internal activations.
\ifCLASSOPTIONcompsoc
  \section*{Acknowledgments}
  This material is based upon work supported by the National Science Foundation under Grant No. 1741431. Any opinions, findings, and conclusions or recommendations expressed in this material are those of the author(s) and do not necessarily reflect the views of the National Science Foundation.
\else
  \section*{Acknowledgment}
\fi

\ifCLASSOPTIONcaptionsoff
  \newpage
\fi

\bibliographystyle{IEEEtran}
\bibliography{references}

\begin{IEEEbiography}[{\includegraphics[width=1in,height=1.25in,clip,keepaspectratio]{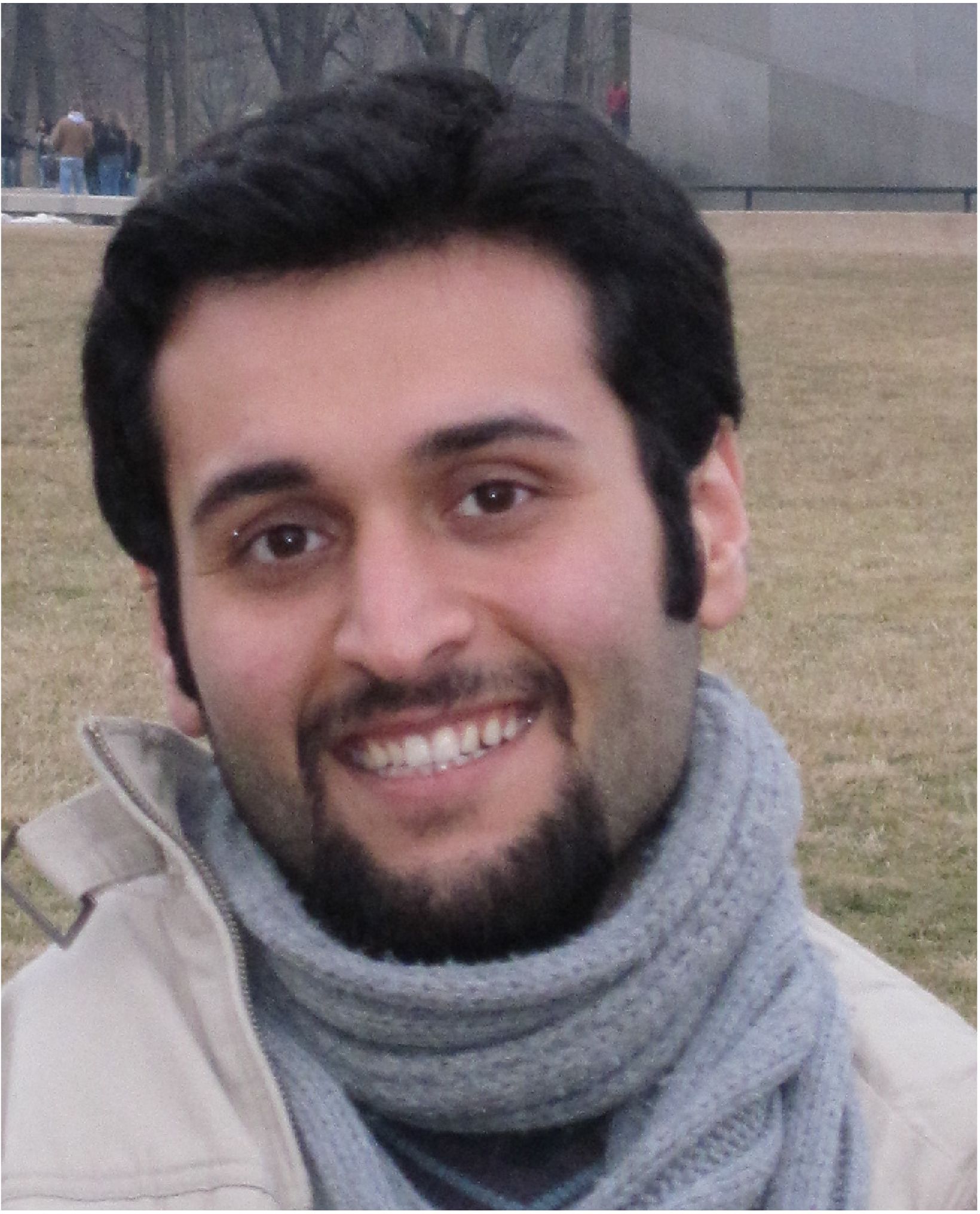}}]{Mahdi M. Kalayeh}
received his B.Sc. from Tehran Polytechnic (Amirkabir University of Technology) in 2009 and M.Sc. from Illinois Institute of Technology (IIT) in 2010, both in Electrical Engineering. At IIT, he was a member of Medical Imaging Research Center (MIRC) where he worked under the supervision of Dr. Jovan G. Brankov on the Generalization Evaluation of Machine Learning Numerical Observers in Image Quality Assessment. From February 2011 to June 2012, Mahdi worked as a visiting researcher at Multimedia Signal Processing Research Laboratory at Electrical Engineering Department of Tehran Polytechnic under the supervision of Dr. Hamid Sheikhzadeh. Since August 2012, Mahdi is a member of Center for Research in Computer Vision (CRCV) at the University of Central Florida where he is currently a Ph.D. candidate in Computer Science and is advised by Dr. Mubarak Shah. His research is on the intersection of Computer Vision and Machine Learning, specifically, it includes Deep Learning, Visual Attribute Prediction, Semantic Segmentation, Complex Event and Action Recognition, Object Recognition and Scene Understanding. Mahdi has published several papers in conferences and journals such as CVPR, ACM Multimedia, IEEE Transactions on Cybernetics, and IEEE Transactions on Medical Imaging. He has also served as a reviewer for peer-reviewed conferences and journals including CVPR, IJCV, IEEE Transactions on Image Processing, and IEEE Transactions on Multimedia.
\end{IEEEbiography}
\vfill
\begin{IEEEbiography}[{\includegraphics[width=1in,height=1.25in,clip,keepaspectratio]{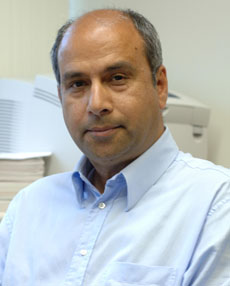}}]{Mubarak Shah}
Mubarak Shah, the Trustee chair professor of computer science, is the founding director of the Center for Research in Computer Vision at the University of Central Florida (UCF). He is an editor of an international book series on video computing, was editor-in-chief of Machine Vision and Applications journal, and an associate editor of ACM Computing Surveys journal. He was the program cochair of CVPR 2008, an associate editor of the IEEE T-PAMI, and a guest editor of the special issue of the International Journal of Computer Vision on Video Computing. His research interests include video surveillance, visual tracking, human activity recognition, visual analysis of crowded scenes, video registration, UAV video analysis, and so on. He is an ACM distinguished speaker. He was an IEEE distinguished visitor speaker for 1997-2000 and received the IEEE Outstanding Engineering Educator Award in 1997. In 2006, he was awarded a Pegasus Professor Award, the highest award at UCF. He received the Harris Corporations Engineering Achievement Award in 1999, TOKTEN awards from UNDP in 1995, 1997, and 2000, Teaching Incentive Program Award in 1995 and 2003, Research Incentive Award in 2003 and 2009, Millionaires Club Awards in 2005 and 2006, University Distinguished Researcher Award in 2007, Honorable mention for the ICCV 2005 Where Am I? Challenge Problem, and was nominated for the Best Paper Award at the ACM Multimedia Conference in 2005. He is a fellow of the IEEE, AAAS, IAPR, and SPIE.
\end{IEEEbiography}

\end{document}